\documentclass[numbers]{article}

\usepackage{natbib}

\usepackage[preprint]{neurips_2024}

\usepackage[utf8]{inputenc} 
\usepackage[T1]{fontenc}    
\usepackage{hyperref}       
\usepackage{url}            
\usepackage{booktabs}       
\usepackage{amsfonts}       
\usepackage{nicefrac}       
\usepackage{microtype}      
\usepackage{xcolor}         

\usepackage{graphicx}
\usepackage{subfigure}
\usepackage{algorithm}
\usepackage{algorithmic}
\usepackage{amsmath}
\usepackage{amssymb}
\usepackage{mathtools}
\usepackage{amsthm}

\usepackage[xindy,acronym,toc,nomain]{glossaries}
\makeglossaries
\usepackage{subcaption}

\theoremstyle{plain}
\newtheorem{theorem}{Theorem}[section]
\newtheorem{proposition}[theorem]{Proposition}

\theoremstyle{definition}

\newtheorem{assumption}[theorem]{Assumption}
\theoremstyle{remark}

\newacronym{UHE}{UHE-BO}{Unknown Hyperparameter Estimation for Bayesian Optimization}
\newacronym{RDEXP3}{RA-BO}{Random Sampling with EXP3 for Bayesian Optimization}
\newacronym{TS}{ts}{Thompson sampling}
\newacronym{MLE}{MLE}{maximum likelihood estimation}
\newacronym{MAP}{MAP}{maximum A Posteriori}
\newacronym{MCMC}{MCMC}{Markov-chain Monte Carlo}
\newacronym{GP}{gp}{Gaussian Process}

\title{Provably Efficient Bayesian Optimization with Unknown Gaussian Process Hyperparameter Estimation}

\author{
Huong Ha  \\RMIT University \\ Australia 
\And
Vu Nguyen  \\Amazon \\ Australia 
\And
Hung Tran-The \\Deakin University \\ Australia
\And
  Hongyu Zhang\\ Chongqing University \\ China 
\And
Xiuzhen Zhang \\ RMIT University \\ Australia
  \And
Anton van den Hengel \\University of Adelaide \\ Australia 
 }

\begin{document}

\maketitle

\global\long\def\se{\hat{\text{se}}}%

\global\long\def\interior{\text{int}}%

\global\long\def\boundary{\text{bd}}%

\global\long\def\new{\text{*}}%

\global\long\def\stir{\text{Stirl}}%

\global\long\def\dist{d}%

\global\long\def\HX{\entro\left(X\right)}%
 
\global\long\def\entropyX{\HX}%

\global\long\def\HY{\entro\left(Y\right)}%
 
\global\long\def\entropyY{\HY}%

\global\long\def\HXY{\entro\left(X,Y\right)}%
 
\global\long\def\entropyXY{\HXY}%

\global\long\def\mutualXY{\mutual\left(X;Y\right)}%
 
\global\long\def\mutinfoXY{\mutualXY}%

\global\long\def\xnew{y}%

\global\long\def\bm{\mathbf{m}}%

\global\long\def\bx{\mathbf{x}}%

\global\long\def\bw{\mathbf{w}}%

\global\long\def\bz{\mathbf{z}}%

\global\long\def\bu{\mathbf{u}}%

\global\long\def\bs{\boldsymbol{s}}%

\global\long\def\bk{\mathbf{k}}%

\global\long\def\bX{\mathbf{X}}%

\global\long\def\tbx{\tilde{\bx}}%

\global\long\def\by{\mathbf{y}}%

\global\long\def\bY{\mathbf{Y}}%

\global\long\def\bZ{\boldsymbol{Z}}%

\global\long\def\bU{\boldsymbol{U}}%

\global\long\def\bv{\boldsymbol{v}}%

\global\long\def\bn{\boldsymbol{n}}%

\global\long\def\bh{\boldsymbol{h}}%

\global\long\def\bV{\boldsymbol{V}}%

\global\long\def\bK{\mathbf{K}}%

\global\long\def\bbeta{\gvt{\beta}}%

\global\long\def\bmu{\gvt{\mu}}%

\global\long\def\btheta{\boldsymbol{\theta}}%

\global\long\def\blambda{\boldsymbol{\lambda}}%

\global\long\def\bgamma{\boldsymbol{\gamma}}%

\global\long\def\bpsi{\boldsymbol{\psi}}%

\global\long\def\bphi{\boldsymbol{\phi}}%

\global\long\def\bpi{\boldsymbol{\pi}}%

\global\long\def\eeta{\boldsymbol{\eta}}%

\global\long\def\bomega{\boldsymbol{\omega}}%

\global\long\def\bepsilon{\boldsymbol{\epsilon}}%

\global\long\def\btau{\boldsymbol{\tau}}%

\global\long\def\bSigma{\gvt{\Sigma}}%

\global\long\def\realset{\mathbb{R}}%

\global\long\def\realn{\realset^{n}}%

\global\long\def\integerset{\mathbb{Z}}%

\global\long\def\natset{\integerset}%

\global\long\def\integer{\integerset}%

\global\long\def\natn{\natset^{n}}%

\global\long\def\rational{\mathbb{Q}}%

\global\long\def\rationaln{\rational^{n}}%

\global\long\def\complexset{\mathbb{C}}%

\global\long\def\comp{\complexset}%

\global\long\def\compl#1{#1^{\text{c}}}%

\global\long\def\and{\cap}%

\global\long\def\compn{\comp^{n}}%

\global\long\def\comb#1#2{\left({#1\atop #2}\right) }%

\global\long\def\nchoosek#1#2{\left({#1\atop #2}\right)}%

\global\long\def\param{\vt w}%

\global\long\def\Param{\Theta}%

\global\long\def\meanparam{\gvt{\mu}}%

\global\long\def\Meanparam{\mathcal{M}}%

\global\long\def\meanmap{\mathbf{m}}%

\global\long\def\logpart{A}%

\global\long\def\simplex{\Delta}%

\global\long\def\simplexn{\simplex^{n}}%

\global\long\def\dirproc{\text{DP}}%

\global\long\def\ggproc{\text{GG}}%

\global\long\def\DP{\text{DP}}%

\global\long\def\ndp{\text{nDP}}%

\global\long\def\hdp{\text{HDP}}%

\global\long\def\gempdf{\text{GEM}}%

\global\long\def\ei{\text{EI}}%

\global\long\def\rfs{\text{RFS}}%

\global\long\def\bernrfs{\text{BernoulliRFS}}%

\global\long\def\poissrfs{\text{PoissonRFS}}%

\global\long\def\grad{\gradient}%
 
\global\long\def\gradient{\nabla}%

\global\long\def\cpr#1#2{\Pr\left(#1\ |\ #2\right)}%

\global\long\def\var{\text{Var}}%

\global\long\def\Var#1{\text{Var}\left[#1\right]}%

\global\long\def\cov{\text{Cov}}%

\global\long\def\Cov#1{\cov\left[ #1 \right]}%

\global\long\def\COV#1#2{\underset{#2}{\cov}\left[ #1 \right]}%

\global\long\def\corr{\text{Corr}}%

\global\long\def\sst{\text{T}}%

\global\long\def\SST{\sst}%

\global\long\def\ess{\mathbb{E}}%

\global\long\def\Ess#1{\ess\left[#1\right]}%

\global\long\def\fisher{\mathcal{F}}%

\global\long\def\bfield{\mathcal{B}}%
 
\global\long\def\borel{\mathcal{B}}%

\global\long\def\bernpdf{\text{Bernoulli}}%

\global\long\def\betapdf{\text{Beta}}%

\global\long\def\dirpdf{\text{Dir}}%

\global\long\def\gammapdf{\text{Gamma}}%

\global\long\def\gaussden#1#2{\text{Normal}\left(#1, #2 \right) }%

\global\long\def\gauss{\mathbf{N}}%

\global\long\def\gausspdf#1#2#3{\text{Normal}\left( #1 \lcabra{#2, #3}\right) }%

\global\long\def\multpdf{\text{Mult}}%

\global\long\def\poiss{\text{Pois}}%

\global\long\def\poissonpdf{\text{Poisson}}%

\global\long\def\pgpdf{\text{PG}}%

\global\long\def\wshpdf{\text{Wish}}%

\global\long\def\iwshpdf{\text{InvWish}}%

\global\long\def\nwpdf{\text{NW}}%

\global\long\def\niwpdf{\text{NIW}}%

\global\long\def\studentpdf{\text{Student}}%

\global\long\def\unipdf{\text{Uni}}%

\global\long\def\transp#1{\transpose{#1}}%
 
\global\long\def\transpose#1{#1^{\mathsf{T}}}%

\global\long\def\mgt{\succ}%

\global\long\def\mge{\succeq}%

\global\long\def\idenmat{\mathbf{I}}%

\global\long\def\trace{\mathrm{tr}}%

\global\long\def\argmax#1{\underset{_{#1}}{\text{argmax}} }%

\global\long\def\argmin#1{\underset{_{#1}}{\text{argmin}\ } }%

\global\long\def\diag{\text{diag}}%

\global\long\def\norm{}%

\global\long\def\spn{\text{span}}%

\global\long\def\vtspace{\mathcal{V}}%

\global\long\def\field{\mathcal{F}}%
 
\global\long\def\ffield{\mathcal{F}}%

\global\long\def\inner#1#2{\left\langle #1,#2\right\rangle }%
 
\global\long\def\iprod#1#2{\inner{#1}{#2}}%

\global\long\def\dprod#1#2{#1 \cdot#2}%

\global\long\def\norm#1{\left\Vert #1\right\Vert }%

\global\long\def\entro{\mathbb{H}}%

\global\long\def\entropy{\mathbb{H}}%

\global\long\def\Entro#1{\entro\left[#1\right]}%

\global\long\def\Entropy#1{\Entro{#1}}%

\global\long\def\mutinfo{\mathbb{I}}%

\global\long\def\relH{\mathit{D}}%

\global\long\def\reldiv#1#2{\relH\left(#1||#2\right)}%

\global\long\def\KL{KL}%

\global\long\def\KLdiv#1#2{\KL\left(#1\parallel#2\right)}%
 
\global\long\def\KLdivergence#1#2{\KL\left(#1\ \parallel\ #2\right)}%

\global\long\def\crossH{\mathcal{C}}%
 
\global\long\def\crossentropy{\mathcal{C}}%

\global\long\def\crossHxy#1#2{\crossentropy\left(#1\parallel#2\right)}%

\global\long\def\breg{\text{BD}}%

\global\long\def\lcabra#1{\left|#1\right.}%

\global\long\def\lbra#1{\lcabra{#1}}%

\global\long\def\rcabra#1{\left.#1\right|}%

\global\long\def\rbra#1{\rcabra{#1}}%

\begin{abstract}
Gaussian process (GP) based Bayesian optimization (BO) is a powerful method for optimizing black-box functions efficiently. The practical performance and theoretical guarantees of this approach depend on having the correct GP hyperparameter values, which are usually unknown in advance and need to be estimated from the observed data. However, in practice, these estimations could be incorrect due to biased data sampling strategies used in BO. This can lead to degraded performance and break the sub-linear global convergence guarantee of BO. To address this issue, we propose a new BO method that can sub-linearly converge to the objective function's global optimum even when the true GP hyperparameters are unknown in advance and need to be estimated from the observed data. Our method uses a multi-armed bandit technique (EXP3) to add random data points to the BO process, and employs a novel training loss function for the GP hyperparameter estimation process that ensures consistent estimation. We further provide theoretical analysis of our proposed method. Finally, we demonstrate empirically that our method outperforms existing approaches on various synthetic and real-world problems.
\end{abstract}

\vspace{-0.3cm}
\section{Introduction} \label{sec:introduction}
\vspace{-0.2cm}
Bayesian optimization (BO) is a powerful technique for optimizing black-box functions, particularly when the function evaluation is expensive and/or when the function's derivative information is not available \cite{Jones_1998Efficient, Brochu_2010Tutorial, Shahriari_2016Taking,cowen2022hebo}. It has been shown to be successful in a variety of tasks that can be cast in terms of function optimization, including new materials discovery \cite{ueno2016combo,li2018accelerating,Hase2021gryffin,Hughes2021MaterialBinding}, drug-target interaction predictions~\cite{ban2017efficient}, environmental monitoring \cite{marchant2012bayesian}, hyperparameter optimization of machine learning models~\cite{Snoek_2012Practical,Hvarfner2022PiBO,parker2022automated}, neural architecture search \cite{kandasamy2018neural, Ru2021InterpretableNAS}, and the life sciences \cite{Cosenza2022BOCellular,Kenan2022LassoBench}.

The key idea of BO is to construct a surrogate model from the observed data, and then use this surrogate model to guide where to sample a next data point. In this way, the function's global optimum can be found in an efficient manner. There are several surrogate models used in the literature~\cite{Snoek_2015Scalable,Bergstra2011TPE, Springenberg_2016Bayesian,TreeBO_ICML2017}, among which Gaussian process (GP) \citep{Rasmussen_2006gaussian} is the most popular due to its analytically tractable for uncertainty estimation. A common assumption of the GP-based BO methods is that the hyperparameters of the GP in which the objective function is drawn from are known in advance, or can be learned accurately from the observed data \citep{Srinivas_2010Gaussian}. This assumption clearly does not always hold, as in practice, one has no information regarding the GP hyperparameters. In the most optimistic scenario, these hyperparameters can only be estimated accurately when the observed data is infinite and independently and identically distributed (i.i.d.). However, in BO, (i) the observed data are normally finite due to limited evaluation budget, and, (ii) the observations are not i.i.d. as BO selects these points using some criteria and based on previous sampled data. It has been shown that incorrect estimations of these hyperparameters can invalidate the sub-linear global convergence guarantee, which is the key advantage of BO over other optimization techniques \cite{powell1973search,back1996evolutionary,hansen2006cma}. 

\begin{figure*}[t]
    \centering
    \includegraphics[trim=0cm 0cm 0cm  0cm, clip, width=0.94\linewidth]{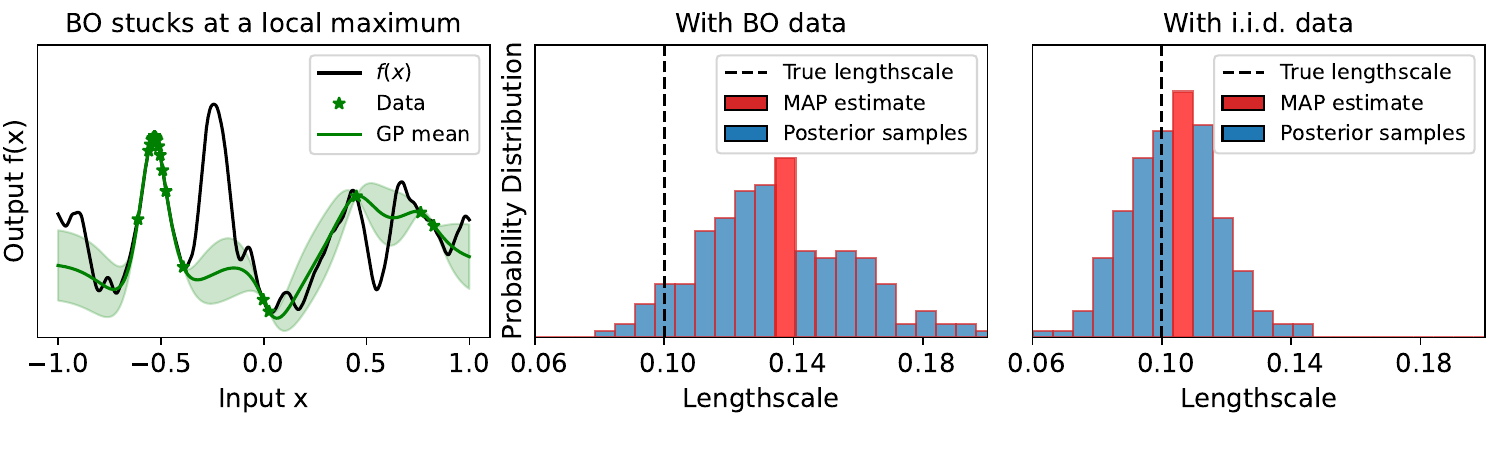}
     \vspace{-14pt}
    \caption{\textbf{Left:} Data selected by a BO process. \textbf{Middle:} Estimated posterior distribution of a GP hyperparameter based on observed data from a BO process. \textbf{Right:} Estimated posterior distribution of the GP hyperparameter based on i.i.d. data (with the same number of data points as in the middle figure). The \textbf{Middle} illustrates that the observed data from a BO process are generally not i.i.d., thus, the estimated GP hyperparameter posterior distribution might not represent the true posterior. The \textbf{Right} shows that even when the GP hyperparameter posterior distribution is estimated from i.i.d. data, the MAP estimate may not be the true GP hyperparameter as the observed data is \textit{finite}. 
    }    
    \label{fig:methodillustration}
    \vspace{-6pt}
\end{figure*}


Limited research has attempted to address the problem of convergence guarantees of BO when the GP hyperparameters are unknown in advance~\cite{Bull_2011Convergence, Wang_2014Theoretical, Berkenkamp2019UnknownHyper, BogunovicK21}. The primary methodologies applied are to reduce the kernel lengthscales over time to expand the function class encoded in the GP models, or to increase the confidence bounds based on the misspecified errors, allowing global convergence. However, the resulting uncertainty can be overestimated by monotonically reducing the lengthscales or increasing the confidence bounds, and thus, such strategies could explore the search domains intensively, decreasing BO efficiency. Furthermore, these methods do not provide accurate estimations of the GP hyperparameters.

In this work, we first investigate the issues in estimating the GP hyperparameters in BO. We then propose a novel approach, namely \underline{U}nknown \underline{H}yperparameter \underline{E}stimation for \underline{B}ayesian \underline{o}ptimization, or \acrshort{UHE}, that can guarantee the sub-linear global convergence when the GP hyperparameters are unknown in advance and need to be estimated from the observed data. The key idea is to construct a consistent loss function for estimating the GP hyperparameters accurately. To do so, first, we construct a  dataset with a sufficient amount of i.i.d. data. 
We then propose a method to approximate the standard GP hyperparameter loss function by a consistent one that can provide asymptotically consistent estimators of the GP hyperparameters. We prove theoretically that \acrshort{UHE} achieves sub-linear global convergence and empirically \acrshort{UHE} outperforms existing state-of-the-art methods on various synthetic and real-world problems. 
 In summary, the contributions of this paper are:
\begin{itemize}
\itemsep0em
    \item We investigate the issues in estimating the hyperparameters of GPs used in BO;
    \item We propose a novel method that can provide consistent GP hyperparameter estimates for BO, addressing the problem of BO with unknown GP hyperparameters;
    \item We provide theoretical analysis guaranteeing the convergence of our proposed method.
\end{itemize}

\vspace{-0.2cm}
\section{Related Work} \label{sec:relatedwork}
\vspace{-0.2cm}
Given a few observations, the performance of BO is very dependent on the accurate estimation of a GP surrogate model. In practice, we typically learn such a GP surrogate model using the maximum likelihood estimation (MLE) or maximum a posteriori (MAP) methods, depending on whether there is prior knowledge about the hyperparameters \cite{Srinivas_2010Gaussian, Shahriari_2016Taking}. The resulting estimates can be incorrect, as (i) there are only a few function evaluations available, and (ii) the training data are non i.i.d. due to BO sampling strategies, as shown in Sec. \ref{sec_issue_estimating_GP_hypers}. These incorrect GP hyperparameter estimates can break the sub-linear global convergence guarantee of BO \cite{Bull_2011Convergence, Wang_2014Theoretical, Berkenkamp2019UnknownHyper, BogunovicK21}. Another approach is to marginalize out the hyperparameters using either quadrature \cite{Osborne2009GPOpt} or Markov Chain Monte Carlo (MCMC) methods~\cite{Snoek_2012Practical, Neal2011HMC}. However, the marginalization still suffers the biased data issue and the estimated posterior distribution is not guaranteed to be accurate. 


The problem of guaranteeing BO convergence in the presence of misspecified or unknown hyperparameters has received little research attention. \citet{Bull_2011Convergence} has derived a regret bound for BO with unknown hyperparameters for the Expected Improvement (EI) acquisition function \cite{Jones_1998Efficient} by introducing lower and upper bounds on the possible range of kernel lengthscales. These regret bounds are obtained for deterministic objective functions when the GP hyperparameters are estimated by the MLE. The work in \citet{Wang_2014Theoretical} extended Bull's result to stochastic setting, but a lower bound on the GP kernel lengthscales needed to be known and pre-specified.~\citet{Berkenkamp2019UnknownHyper} develop a GP-UCB based method that guaranteed the sub-linear convergence rate without any knowledge of the GP hyperparameters. The idea is to reduce the GP kernel lengthscales over time to expand the function class encoded in the GP models. However, this method's performance is conservative as reducing the kernel lengthscales monotonically over time can lead to intensive exploration of the search domain. We show empirically (Sec.~\ref{sec:experiment}) that our method, \acrshort{UHE}, outperforms this technique significantly. Recently, the work in \citet{BogunovicK21} addressed the misspecified kernel problem by adjusting the confidence bounds based on the misspecified errors to guarantee global convergence. However, it is unclear how these methods perform empirically as no experiments were conducted. Other BO algorithms such as \cite{Srinivas_2010Gaussian,Bogunovic_2016Truncated,bogunovic2016time,scarlett2018tight,PB2,cai2021lenient} provide theoretical global convergence guarantees asymptotically, but only when GP hyperparameters are known in advance or can be learned accurately from the observed data, which may be impractical. 

Other works relevant to our proposed method are as follows. \citet{Malkomes2018AutomatedBO} suggest creating a space of GP models that is general enough to explain most of datasets, and then selecting an appropriate GP model from this model space. \citet{Farquhar2021BiasedAL} propose a new technique to correct the prediction loss function due to a biased data sampling strategy in active learning. However, this work is only applicable to pool-based active learning where the goal is to train a prediction model from a pool of finite data. In contrast, our work is focused on BO when the domain is continuous (infinite data). The work in~\citet{Hvarfner2023BOwAL} considers misspecified GP hyperparameters and introduces a new acquisition function to correct this issue. Unlike our work, this work focuses on developing a new acquisition function, and does not guarantee global convergence. Our work can be complementary to this work by providing accurate posterior distributions of GP hyperparameters that can be used with the acquisition function proposed in this work. The work in \cite{fan2024transfer} can provide consistent GP hyperparameter estimates but it works in the transfer learning setting, requiring historical data.

In a different line of research, \citet{Hoffman2011Portfolio} propose to use a MAB algorithm (e.g., Hedge~\cite{Auer2022Hedge} or EXP3 \cite{Auer2002nonstochastic}) in BO, to choose the best acquisition function from a set of multiple acquisitions. The reward is defined using the GP prediction. Our method is different from \citet{Hoffman2011Portfolio} in that we use MAB to determine when to sample i.i.d. data points to address the GP hyperparameter estimation issue caused by the biased data sampling strategy of BO to learn these hyperparameters accurately.

\vspace{-0.1cm}
\section{Problem Statement}
\vspace{-0.2cm}

Bayesian optimization (BO) aims at finding the global optimum $x^*$ of an unknown objective function $f:\mathbb{R}^d \mapsto \mathbb{R}$ using a minimal number of evaluations \cite{Jones_2001Taxonomy,Brochu_2010Tutorial,Shahriari_2016Taking,frazier2018tutorial},
\begin{equation} \label{eq:bo}
x^* = {\operatorname{argmax}}_{x \in \mathcal{X}} \ f(x),
\end{equation}
where $ \mathcal{X} \subset \mathbb{R}^d$  is a compact domain. It is generally assumed that we have no access to the gradient of $f(\cdot)$, and the evaluations of $f(\cdot)$ are corrupted by Gaussian noise $\epsilon \sim \mathcal{N}(0, \sigma^2)$. BO solves the problem in Eq. (\ref{eq:bo}) sequentially. Details about BO are in App. \ref{sec:app-bo}.

This work focuses on a popular setting of BO wherein a GP \cite{Rasmussen_2006gaussian} is used as the surrogate model and the GP hyperparameters are unknown in advance and need to be estimated from observed data. We investigate the issues in estimating the GP hyperparameters in BO, and propose a method to provide consistent GP hyperparameter estimates, guaranteeing BO's sub-linear global convergence.



\vspace{-0.1cm}
\section{Issues in Estimating GP Hyperparameters for Bayesian Optimization} \label{sec_issue_estimating_GP_hypers}
\vspace{-0.2cm}

\subsection{Inconsistency Issue of GP Hyperparameter Estimation for Bayesian Optimization} \label{sec:issue-bias}
\vspace{-0.1cm}
Let us denote a GP model using a hyperparameter $\theta$ as $\mathcal{GP} \bigl(\mu_{\theta}(x), k_{\theta}(x,x') \bigr)$. At iteration $t$, given the observed data $D_{t-1}=\{x_i, y_i \}_{i=1}^{\vert D_{t-1} \vert}$, the standard practice of estimating the hyperparameter $\theta$ is to find the value $\hat{\theta}_t$ that minimizes a loss function $\mathcal{L} \bigl(\by_{t-1}, \bx_{t-1}, \theta \bigr)$ where $\bx_{t-1}=[x_1, \dots,x_{\vert D_{t-1} \vert}]^T$, $\by_{t-1}=[y_1, \dots,y_{\vert D_{t-1} \vert}]^T$. Some common choices of $\mathcal{L} \bigl(\by_{t-1}, \bx_{t-1}, \theta \bigr)$ include the log marginal likelihood and the log posterior distribution function, resulting in the Maximum Likelihood Estimation (MLE) and Maximum A Posteriori (MAP) estimates \cite{Rasmussen_2006gaussian} (see App. \ref{app_sec_GP_LogMarginalLLK}).


In traditional supervised learning, the estimation obtained when minimizing a loss function can be a consistent estimate of the true value when the data used to construct this loss function are drawn i.i.d. from the population data distribution \cite{Farquhar2021BiasedAL}. For GP, it has been shown that, for various kernels, when the observed data $\{x_i\}_{i=1}^{\vert D_{t-1} \vert}$ are drawn i.i.d. from the population data distribution, the MLE or MAP estimate can be a consistent estimate of the true hyperparameter \cite{Bachoc2013MLE}. However, this is not guaranteed when the observed data are not i.i.d. This is particularly the case in BO as the observed data at an iteration $t$ are generally not i.i.d. because the BO algorithm conditions on the previous observed data $D_{t-1}$ to select the next data point $x_t$. Therefore, the estimate (e.g., MAP or MLE) $\hat{\theta}_t$ obtained may not be a consistent estimate of the true hyperparameter $\theta^*$, even when $\vert D_t \vert \to \infty$. It is worth noting that this issue is not limited to BO, but also occurs in most active learning based methods~\cite{Farquhar2021BiasedAL}.


\vspace{-0.1cm}
\subsection{Uncertainty in GP Hyperparameter Estimation for Bayesian Optimization}
\vspace{-0.1cm}
As discussed in Sec. \ref{sec:issue-bias}, when $t \rightarrow \infty$, if the observed data $\{x_i\}_{i=1}^{\vert D_{t-1} \vert}$ are drawn i.i.d. from the population data distribution, the estimate $\hat{\theta}_t$ obtained by minimizing the loss function $\mathcal{L}\bigl(\by_{t-1}, \bx_{t-1}, \theta \bigr)$ may converge to the true $\theta^*$. However, when $t$ is finite, $\hat{\theta}_t$ may not be $\theta^*$ any more. 


We illustrate in Fig. \ref{fig:methodillustration} the inconsistency and uncertainty issues in GP hyperparameter estimation of BO with a synthetic 1-dim function. The left plot shows that the observed data obtained during a BO process may not be i.i.d. The middle plot is the GP hyperparameter posterior distribution estimated from the non i.i.d. data obtained by BO that might not represent the true posterior distribution. The right plot shows a GP hyperparameter posterior distribution constructed from i.i.d. data. Even in this ideal case with i.i.d. data, the MAP estimate may still not be the true GP hyperparameter as the observed data are finite. There exists a high probability of the true hyperparameter being different.

Existing BO methods and their theoretical analysis \cite{Srinivas_2010Gaussian,Bogunovic_2016Truncated,bogunovic2016time,Nguyen_ACML2017Regret,scarlett2018tight,PB2,cai2021lenient} generally assume the GP hyperparameters are known prior to the optimization process (except those we described in Secs.~\ref{sec:introduction} and \ref{sec:relatedwork}). Therefore, they are vulnerable to inaccurate estimation of the GP hyperparameters. When using these methods, the global convergence guarantee of BO can be invalidated, and thus, it might not be able to achieve the convergence rate that was reported.


\vspace{-0.1cm}
\section{Consistent GP Hyperparameter Estimation for Bayesian Optimization} \label{sec:method}
\vspace{-0.2cm}
We propose a novel approach to tackle the inconsistency issue in GP hyperparameter estimation of BO. Our main idea is to construct a consistent loss function when estimating the GP hyperparameters. To achieve this, we first develop a method to efficiently sample i.i.d. data during the BO process (Sec. \ref{sec:samplingiid}), and then a method to approximate the original inconsistent loss function by a consistent one to achieve consistent GP hyperparameter estimation (Sec. \ref{sec:unbiasedloss}).

\begin{algorithm}[tb] 
   \caption{Unknown Hyperparameter Estimation for Bayesian Optimization (\acrshort{UHE})}
   \label{alg:new}
\begin{algorithmic}[1]
   \STATE {\bfseries Input:} initial data $D_0$,  budget evaluation $T$, GP loss function $\mathcal{L}(.)$, choice of $M_t$
   \STATE {\bfseries Output:} An estimate $\hat{x}^*$ for the global optimum $x^*$
   \STATE Initialize $\omega^m_0=1$ for $m=1,2$
   \FOR {$t = 1, \dots, T$}
   \STATE 
   $\gamma=\sqrt{4\log2/((e-1)T)}$
   \IF {$t$ is odd}   
      \STATE Set $p^m_t= \frac{ (1-\gamma) \omega^m_{t-1} }{\sum_{m=1}^2 \omega^m_{t-1}} + \frac{\gamma}{2}$ for $m=1,2$
     and pull an arm $a_t \in \{1,2\} \sim [p^1_t, p^2_t]$
 \ELSE
    \STATE $a_t = a_{t-1}$
   \ENDIF
   \IF {$a_t = 1$ \textbf{and} $t$ is odd}
        \STATE Select a random point $x_t \in \mathcal{X}$, evaluate $y_t = f(x_t)+\epsilon_t$, and update $D_t = D_{t-1} \cup \{x_t, y_t\}$
    \ELSE
       \STATE Sample $\{x'_j\}_{j=1}^{M_t} \in \mathcal{X}$ randomly and find $\hat{x}'_j = {\operatorname{argmin}}_{x \in D_{t-1}} \Vert x - x'_j \Vert_2$ for each $x'_j$
       \STATE Find $\hat{\theta}_t = {\operatorname{argmin}}_{\theta} \ \mathcal{L} (.)$ in Eq. (\ref{eq:loss-function})
        and update a GP using $\hat{\theta}_t$ and $D_{t-1}$  in Eq. (\ref{eq:new-GP})
        \STATE Find $x_t = \operatorname{argmax}_{x\in \mathcal{X}} \alpha(x; D_{t-1})$, evaluate $y_t= f(x_t)+\epsilon_t$, update $D_t = D_{t-1} \cup \{x_t, y_t\}$
   \ENDIF
   \IF {$t$ is even} 
   \STATE Compute the reward $r_t^{a_{t-1}} = \max(y_t, y_{t-1})$ and scale it using $D_0$
   \STATE For $m=1,2$, set $\omega^m_{t} = \omega^m_{t-2} \text{exp} \bigl(\gamma \frac{r^m_t}{2 \times p^m_{t-1}} \bigr)$ if $m==a_{t-1}$ and $\omega^m_{t}=\omega^m_{t-2}$ otherwise 
   \ENDIF
   \ENDFOR
\end{algorithmic}
\end{algorithm}

\vspace{-0.1cm}
\subsection{EXP3-based Algorithm for Efficient Sampling of I.I.D. Samples} \label{sec:samplingiid}


A straightforward approach to obtain consistent estimates of the GP hyperparameters is to sample i.i.d. data points during the BO process and use them to construct the loss function $\mathcal{L} \bigl(\by_{t-1}, \bx_{t-1}, \theta \bigr)$. However, this approach can degrade BO performance significantly as these i.i.d. data points might not provide the most valuable information for determining the global optimum of $f(\cdot)$. Thus, it is critical to decide when to sample i.i.d. data points to improve the GP hyperparameters learning and when to sample informative data points to learn about the global optimum's location. 

In this work, we formulate this task as a multi-armed bandit (MAB) problem with two arms \cite{Auer2002finite}. The first arm is to sample a random data point to construct an i.i.d. dataset to address the inconsistency issue in the GP hyperparameter estimation process, enabling accurate estimation. The second arm is to sample the data point recommended by the acquisition function assuming the correct GP modeling, thus improving the understanding of the global optimum's location. We propose to use EXP3 \cite{Auer2002nonstochastic} to select the optimal arm in each iteration $t$. To do so, EXP3 associates each arm $m$ with a reward $r^m_t$, and the goal is to select an arm $a_t \in \{1,2\}$ in each iteration $t$ to maximize the total reward $\sum_{t=1}^T r^{a_t}_t$ in the long run. The probability $p_t^m$ of choosing each arm $m$ at iteration $t$ is computed based on its cumulative reward so far (see \cite{Auer2002nonstochastic}). After choosing an arm $a_t$, the algorithm receives the reward $r^{a_t}_t$, and the probabilities $\{p_t^m\}_{m=1}^2$ are updated accordingly. The rationale behind our choice of EXP3 is that: (i) it applies to the partial information setting, i.e., in each iteration, it only requires knowing the reward of one arm whilst other MAB algorithms might require the rewards of all arms, and (ii) it achieves logarithmic expected regret without assuming the i.i.d. or Gaussian assumption on the reward generating distribution. More information about EXP3 is in App. \ref{sec:app-exp3}.

 In our proposed method, at each iteration $t$, we compute the reward $r^m_t$ for each arm $m \in \{1,2\}$ as follows. For the first arm ($m=1$), we sample two data points: the first one $x_{t,1}$ via random sampling and the second one $x_{t,2}$ via the acquisition function constructed based on the GP model using the existing observed dataset $D_{t-1}$ and the first data point $\{x_{t,1}, y_{t,1} \}$. For the second arm ($m=2$), we sample two data points via the acquisition function to ensure the reward comparable with the first arm. The reward of these arms is defined as the maximum objective function value of these two data points, i.e., $r^m_t = \max(y_{t,1}, y_{t,2}), m=1,2$. The rationale behind this design of reward is to identify how much performance we can gain if we aim to estimate the GP hyperparameters more accurately (by sampling i.i.d. data points). When arm $m=1$ achieves higher reward than arm $m=2$, it means we should continue to sample i.i.d. data points so as to learn the GP hyperparameters better.

\vspace{-0.1cm}
\subsection{Consistent GP Hyperparameter Estimation Loss Function} \label{sec:unbiasedloss}
\vspace{-0.1cm}

 With the sampling methodology proposed in Sec. \ref{sec:samplingiid}, the collected data points are mixed between (i) sampled randomly and (ii) based on the acquisition function.
 Therefore, the observed data are not completely i.i.d. and the GP hyperparameter estimate is still inconsistent. To completely correct this inconsistency issue, we propose to approximate the inconsistent GP hyperparameter loss function by a consistent one. Our idea is to sample $M_t$ random data points $ \{x'_j \}_{j=1}^{M_t} \in \mathcal{X}$ ($M_t \geq \vert D_{t-1} \vert$), then approximate the objective function values of these data points by the closest data points in the observed dataset whose objective function values are known, i.e., $\hat{x}'_j = {\operatorname{argmin}}_{x \in D_{t-1}} \Vert x - x'_j \Vert_2$, and $\Vert . \Vert_2$ is the L2 norm. The new GP hyperparameter loss function is constructed from these i.i.d. data points and the GP hyperparameter estimate is obtained by minimizing this new loss function as, 
 \begin{equation} \label{eq:loss-function}
 \begin{aligned}
   \hat{\theta}_t = {\operatorname{argmin}}_{\theta} \  \mathcal{L} \bigl(\hat{\by}_{t-1}', \bx_{t-1}, \theta \bigr),
 \end{aligned}
 \end{equation}
 where $\bx_{t-1}'=[x'_1,...,x'_{M_t}]^T$, $\hat{\by}_{t-1}'=[y(\hat{x}'_1),...,y(\hat{x}'_{M_t})]$.

The motivation of this new GP hyperparameter loss function is that, given the common assumption of the objective function $f(.)$ is a sample path from a GP with a bounded Lipschitz constant, the function values of two data points close to each other are close to each other. Thus, when the observed data increases, the distance between two randomly sampled data points decreases, and the approximation accuracy in Eq. (\ref{eq:loss-function}) also increases. In Sec. \ref{sec:theory}, we prove that for some common loss functions and kernels, the estimates obtained by our proposed loss functions are consistent with high probability.

Note that this new training loss function is only used for estimating the GP hyperparameters. After obtaining an estimate $\hat{\theta}_t$, to compute the GP predictive mean and standard deviation for a new data point $x_{\text{n}}$, we use the original observed dataset $D_{t-1}$, i.e., $\mu\left(x_{\text{n}}\right) = \mathbf{k}_{\hat{\theta}_t, \text{n}} \big[ \bK_{\hat{\theta}_t} + \sigma^2 \idenmat_{t-1} \big]^{-1}\mathbf{y}_{t-1}$, $\sigma^{2}\left(x_{\text{n}}\right) = k_{\hat{\theta}_t}(x_{\text{n}}, x_{\text{n}})-\mathbf{k}_{\hat{\theta}_t, \text{n}} \big[ \bK_{\hat{\theta}_t} + \sigma^2 \idenmat_{t-1} \big]^{-1} \mathbf{k}_{\hat{\theta}_t,\text{n}}^{T}$, $\by_{t-1}=[y_1, ...,y_{\vert D_{t-1} \vert}]^T$, $\bK_{\hat{\theta}_t}=[k_{\hat{\theta}_t}(x_i, x_j)]_{i,j=1}^{\vert D_{t-1} \vert}$, and $\idenmat_{t-1}$ is the $\vert D_{t-1} \vert \times \vert D_{t-1} \vert$ identity matrix.


A pseudocode of our proposed method, \acrshort{UHE}, is summarized in Algorithm \ref{alg:new}.



\vspace{-0.1cm}
\section{Theoretical Analysis for UHE-BO} \label{sec:theory}
\vspace{-0.2cm}

In this section, we derive theoretically the upper bound on the regret of our method, \acrshort{UHE}. Due to the space limit, we state the main theoretical claims and refer to the App. \ref{appendix_sec:theory} for the proofs. We first list the assumptions used to develop our theoretical analysis.


\begin{assumption} \label{assum:l-loss}
    The loss function $\mathcal{L}$ is additive in observations, i.e., $\mathcal{L}(\by, \bx, \theta)= \sum_{i=1}^N l(y_i, x_i, \theta)$ with $\by=[y_1,\dots,y_N], \bx=[x_1,\dots,x_N]$. The function $l$, $\partial l / \partial \theta $, and $\partial^2 l / \partial \theta^2 $ are Lipschitz continuous on $f$ with high probability. That is, $\exists p, p^\prime, p^{\prime\prime}, q, q^\prime, q^{\prime\prime}: \forall x_i, x_j \in \mathcal{X}$, we have that, w.p. $\ge 1-pe^{-(C_l/q)^2}$, $\ge 1-p^\prime e^{-(C_{l^\prime}/q^\prime)^2}$, and $\ge 1-p^{\prime\prime} e^{-(C_{l^{\prime\prime}}/q^{\prime\prime})^2}$, respectively: $| l ( y(x_i), x_i, \theta ) - l( y(x_j), x_i, \theta ) | \le C_l ||  f(x_i) - f(x_j)  ||_1$, $| \partial l (y(x_i), x_i, \theta )/\partial \theta - \partial l ( y(x_j), x_i, \theta )/\partial \theta | \le C_{l^\prime} ||  f(x_i) - f(x_j)  ||_1$, and, $| \partial^2 l (y(x_i), x_i, \theta )/\partial \theta^2 - \partial^2 l ( y(x_j), x_i, \theta )/\partial \theta^2 | \le C_{l^{\prime\prime}} ||  f(x_i) - f(x_j)  ||_1$.
\end{assumption}

\begin{assumption} \label{assum:f-smoothness}
Let $\mathcal{X} \subset [0, r]^d$ be compact and convex, $d\in \mathbb{N}, r>0$. The true kernel $k_{\theta^*}(x, x')$ satisfies the following high probability bound on the derivatives of GP sample paths $f$: there exist some constants $a,b>0$ that $\text{Pr}\{ \sup_{\bx\in \mathcal{X}} \vert \partial f/\partial x_h \vert > L\} \leq ae^{-(L/b)^2}, h=1,\dots,d$.
\end{assumption}

\begin{assumption} \label{assum:Bernstein-von-Mises}
The GP prior mean is $0$. The prior distribution $p(\theta)$ of the GP hyperparameter $\theta$ is continuous and positive in an open neighborhood of the true parameter $\theta^*$. Let us denote $\bx_N'$ as $N$ i.i.d. data points sampled from the search domain $\mathcal{X}$ and $\by_N'$ as the noisy function values of $\bx_N'$, then the GP hyperparameter loss function $\mathcal{L}$ satisfies that $\partial \mathcal{L}(\by_N', \bx_N', \theta ) / \partial \theta$ and $\partial^2 \mathcal{L} \bigl(\by_N', \bx_N', \theta \bigr) / \partial \theta^2$ exist and are continuous in $\theta$. The Fisher information matrix $I(\theta) = -\mathbb{E}_{\theta} [\partial^2 \mathcal{L} \bigl(\by_N', \bx_N', \theta \bigr) / \partial \theta^2]$ is continuous, positive, and upper bounded. The estimate $\hat{\theta}_N={\operatorname{argmin}}_{\theta} \mathcal{L} \bigl(\by_N', \bx_N', \theta \bigr)$ is consistent.
\end{assumption}

 Assumption \ref{assum:l-loss} is a standard assumption in analyzing the consistency property of machine learning model parameters \cite{Farrell2021}. Assumption \ref{assum:f-smoothness} is a commonly used in analyzing regret of BO algorithms \cite{Srinivas_2010Gaussian, bogunovic2016time, PB2}. Assumption \ref{assum:Bernstein-von-Mises} is usually used for the Bernstein-von Mises theorem \cite{Borwanker1971BvMTheorem, Vaart1998AsymStats} which provides convergence analysis for Bayesian parametric models. The assumption regarding the prior distribution of the GP hyperparameters are applicable for various common kernels such as the Square Exponential kernel and the Mat\'{e}rn kernel. The assumption regarding the property of the loss function $\mathcal{L}$ is applicable for various widely-used loss functions such as the log posterior distribution function and the log marginal likelihood function. Finally, the assumption of the estimate $\hat{\theta}_N$ being consistent when the training data $\bx_N'$ are i.i.d. is a standard assumption in traditional supervised learning \cite{Farquhar2021BiasedAL}.


First, we derive an upper bound on the regret of the BO algorithm of sequentially selecting one data point via random sampling and one via the acquisition function (i.e., the first arm in our EXP3-based sampling strategy described in Sec. \ref{sec:samplingiid}). In the sequel, we refer to this algorithm as \acrshort{RDEXP3}.



\vspace{0.2cm}
\begin{proposition} \label{prop:newloss}
 Suppose Assumptions \ref{assum:l-loss} \& \ref{assum:f-smoothness} are satisfied, with $M_t \geq \vert D_t \vert$, then  \acrshort{RDEXP3} achieves, 
\begin{enumerate}
\itemsep=-0.2em

\item With probability $>(1 - dae^{-(L/b)^2} - pe^{-(C_l/q)^2})(1 - t^{1/2}(1-1/t^{1/2}) ^{\lfloor (t-1)/2 \rfloor})(1 - t^{1/2}(1-1/t^{1/2}) ^{M_t})$, $ \vert  \mathcal{L} (\hat{\mathbf{y}}^\prime_t, \mathbf{x}^\prime_t, \theta) - \mathcal{L} (\mathbf{y}^\prime_t, \mathbf{x}^\prime_t, \theta) \vert \leq \mathcal{O}(L C_l t^{-1/2d})$, where $\by_t'$ denotes the noisy function values of $\bx_t'$ and $\hat{\mathbf{y}}^\prime_t$ is defined in Eq. (\ref{eq:loss-function}).
    
\item If the estimate obtained by the true loss function $\mathcal{L} \bigl(\by', \bx', \theta \bigr)$ converges to the true hyperparameter $\theta^*$ then there exist $T_1$ such that for all iteration $t > T_1$, the estimate obtained by the proposed loss function $\mathcal{L} \bigl(\hat{\by}', \bx', \theta \bigr)$ as in Eq. (\ref{eq:loss-function}) also converges to $\theta^*$ with probability $(1 - dae^{-(L/b)^2} - pe^{-(C_l/q)^2} - p^\prime e^{-(C_{l^\prime}/q^\prime)^2}) \prod_{i=T_1}^t(1 - i^{1/2}(1-1/i^{1/2}) ^{\lfloor (i-1)/2 \rfloor}) \prod_{i=T_1}^t (1 - i^{1/2}(1-1/i^{1/2}) ^{M_i})$.
\end{enumerate}
\end{proposition}


\vspace{-0.05cm}
Note the probability in Bullet 2 of Proposition \ref{prop:newloss} is close to $1$ when $t, T_1, L, C_l, C''_l$ are large (App. \ref{sec:app-lest}). Proposition \ref{prop:newloss} guarantees our proposed GP hyperparameter loss function in Eq. (\ref{eq:loss-function}) based on pseudo function values shares the same convergence property with the true loss function w.h.p.


\vspace{0.1cm}
\begin{theorem} \label{theorem:thetaconvergence}
Suppose Assumptions \ref{assum:l-loss}, \ref{assum:f-smoothness}, and \ref{assum:Bernstein-von-Mises} are satisfied. Let us denote $\hat{\theta}_t^{\text{\acrshort{RDEXP3}}}$ as the hyperparameter estimate obtained at iteration $t$ of \acrshort{RDEXP3}. There exists $T_1$ such that for all iteration $t > T_1$, with probability $>(1 - dae^{-(L/b)^2} - pe^{-(C_l/q)^2} - p^\prime e^{-(C_{l^\prime}/q^\prime)^2}) \prod_{i=T_1}^t(1 - i^{1/2}(1-1/i^{1/2}) ^{\lfloor (i-1)/2 \rfloor}) \prod_{i=T_1}^t (1 - i^{1/2}(1-1/i^{1/2}) ^{M_i})$, when $t \rightarrow \infty$, $\sqrt{t}(\hat{\theta}_t^{\text{\acrshort{RDEXP3}}}-\theta^*)\xrightarrow {} 0$. This means, with this probability, $\hat{\theta}_t^{\text{\acrshort{RDEXP3}}} = \theta^*+\mathcal{O} (t^{-1/2})$ for all $t>T_1$.
\end{theorem}

Theorem \ref{theorem:thetaconvergence} directly applies the Bernstein-von Mises theorem \cite{Borwanker1971BvMTheorem, Vaart1998AsymStats}, providing a convergence rate for the hyperparameter estimate obtained by \acrshort{RDEXP3} w.h.p. Let denote the simple regret of \acrshort{RDEXP3} as $S^{\text{\acrshort{RDEXP3}}}_T = f(x^*)- \max_{t=1,..,T} f(x^{\text{\acrshort{RDEXP3}}}_t)$ where $x^{\text{\acrshort{RDEXP3}}}_t$ is the point sampled at iteration $t$ of \acrshort{RDEXP3}. We derive an upper bound for $S^{\text{\acrshort{RDEXP3}}}_T$ when GP-UCB is used as the acquisition function.


\begin{theorem} \label{theorem:regret-random}
Suppose Assumptions \ref{assum:l-loss}, \ref{assum:f-smoothness}, and \ref{assum:Bernstein-von-Mises} are satisfied. Let us define $\gamma_T(\theta) = \max_{A \subset \mathcal{X}: \vert A \vert = T} 0.5 \log \vert \mathbf{I}_{\vert A \vert} + \sigma^{-2} [k_{\theta}(x,x')_{x,x'\in A}] \vert$ as the maximum information gain achieved by sampling $T$ data points in a GP with kernel $k_{\theta}(x,x')$ over $\mathcal{X}$. Then,
\vspace{-0.2cm}
\begin{enumerate}
\itemsep=-0.2em
    \item With probability $>(1 - dae^{-(L/b)^2} - pe^{-(C_l/q)^2} - p^\prime e^{-(C_{l^\prime}/q^\prime)^2}) \prod_{i=T_1}^t(1 - i^{1/2}(1-1/i^{1/2}) ^{\lfloor (i-1)/2 \rfloor}) \prod_{i=T_1}^t (1 - i^{1/2}(1-1/i^{1/2}) ^{M_i})$, there exists $T_2\ (T_2 > T_1)$ that $\forall t>T_2$,  the estimate $\hat{\theta}_t^{\text{\acrshort{RDEXP3}}}$ will satisfy the requirements that for all sample paths $f'$ drawn from the GP with kernel $k_{\hat{\theta}_t^{\text{\acrshort{RDEXP3}}}}(x,x')$, there exist positive constants $L_{t}, a_{t}, b_{t}$ that $\text{Pr}\{ \sup_{\bx\in \mathcal{X}} \vert \partial f'/\partial x_h \vert > L_{t}\} \leq a_{t}e^{-(L_{t}/b_{t})^2}, h=1,\dots,d$.
    \item With GP-UCB as the acquisition function, $\delta \in (0,1)$, and $\beta_t = 2\log \bigl((t/2)^22\pi^2/(3\delta) \bigr) + 2d\log \bigl((t/2)^2 db_{t}r\sqrt{\log(4da_{t}/\delta)} \bigr)$, then $\forall T \! > \! T_2$, $S^{\text{\acrshort{RDEXP3}}}_T \leq \mathcal{O} \Bigl( \sqrt{2\beta_{T} \gamma_{T/2}(\hat{\theta}_T^{\text{\acrshort{RDEXP3}}})/T} \Bigr)$ with probability larger than $(1 - dae^{-(L/b)^2} - pe^{-(C_l/q)^2} - p^\prime e^{-(C_{l^\prime}/q^\prime)^2} - p^{\prime\prime} e^{-(C_{l^{\prime\prime}}/q^{\prime\prime})^2}) \prod_{i=T_2}^T (1 - i^{1/2}(1-1/i^{1/2}) ^{\lfloor (i-1)/2 \rfloor})(1 - i^{1/2}(1-1/i^{1/2}) ^{M_i})$.
\end{enumerate}
\vspace{-0.2cm}
\end{theorem}

The regret bound in Theorem \ref{theorem:regret-random} depends on the maximum information gain $\gamma_{T/2}(\hat{\theta}_T^{\text{\acrshort{RDEXP3}}})$. In the proposition below, we provide the upper bounds on this maximum information gain for various common kernels. This shows, w.h.p., the simple regret of \acrshort{RDEXP3} has a sub-linear rate with large $T$.

\begin{proposition} \label{prop:informationgain}
 Suppose Assumptions \ref{assum:l-loss}, \ref{assum:f-smoothness}, and \ref{assum:Bernstein-von-Mises} are satisfied. For all $T>T_1$ with $T_1$ defined as in Theorem \ref{theorem:thetaconvergence}, with probability $>(1 - dae^{-(L/b)^2} - pe^{-(C_l/q)^2} - p^\prime e^{-(C_{l^\prime}/q^\prime)^2} - p^{\prime\prime} e^{-(C_{l^{\prime\prime}}/q^{\prime\prime})^2}) \prod_{i=T_1}^T (1 - i^{1/2}(1-1/i^{1/2}) ^{\lfloor (i-1)/2 \rfloor})(1 - i^{1/2}(1-1/i^{1/2}) ^{M_i})$, for the Square Exponential kernel, we have that $\gamma_{T/2}(\hat{\theta}_T^{\text{\acrshort{RDEXP3}}}) = \mathcal{O} ((\log (T/2))^{d+1})$, and for the Mat\'{e}rn kernel with the $\nu$ hyperparameter, we have that, $\gamma_{ T/2}(\hat{\theta}_T^{\text{\acrshort{RDEXP3}}}) = \mathcal{O} \bigl((T/2)^{d(d+1)/(2\nu+d(d+1))} \log (T/2)\bigr)$. 
\end{proposition}

Finally, we derive an upper bound of the expected simple regret of our proposed method \acrshort{UHE},  $\mathbb{E}[S_T] = f(x^*) - \mathbb{E} \bigl[\max_{t=1,\dots,T} f(x_t) \bigr] $, when GP-UCB is used as the acquisition function \cite{Srinivas_2010Gaussian}. Note here the expectation is due to the randomness of the EXP3 algorithm. 





\begin{theorem} \label{theorem:final-convergence}
Using GP-UCB as the acquisition function and suppose Assumptions \ref{assum:l-loss}, \ref{assum:f-smoothness}, and \ref{assum:Bernstein-von-Mises} hold. With large $T>T_2$, we have with probability $>(1 - dae^{-(L/b)^2} - pe^{-(C_l/q)^2} - p^\prime e^{-(C_{l^\prime}/q^\prime)^2} - p^{\prime\prime} e^{-(C_{l^{\prime\prime}}/q^{\prime\prime})^2}) \prod_{i=T_2}^T (1 - i^{1/2}(1-1/i^{1/2}) ^{\lfloor (i-1)/2 \rfloor})(1 - i^{1/2}(1-1/i^{1/2}) ^{M_i})$ that our expected simple regret $\mathbb{E}[S_T]$
is upper bounded by,
\begin{enumerate}
\itemsep=-0.2em
    \item $\mathcal{O} \big(\sqrt{2\log 2/T} \! + \! \sqrt{2\beta_{T} ((\log (T/2))^{d+1})/T}\big)$ for SE kernel,
    \item $\mathcal{O} \big(\! \sqrt{2\beta_{T} ((T/2)^{d(d+1)/(2\nu+d(d+1))} \log (T/2))/T} + \sqrt{2\log 2/T} \! \big)$ for Mat\'{e}rn kernel.
\end{enumerate}
\end{theorem}

\paragraph{Discussion.} Note similar to the discussion under Proposition \ref{prop:newloss}, the probability in Theorem \ref{theorem:final-convergence} is close to $1$ when $T, T_2, L, C_l, C''_l$ are large (App. \ref{sec:app-lest}). Compared to the existing regret upper bound for GP-UCB \citep{Srinivas_2010Gaussian} when the true GP hyperparameters are known, our bound in Theorem \ref{theorem:final-convergence} focuses on the simple regret, applies to large $T$ values w.h.p, and has an additional term $\mathcal{O} (\sqrt{2\log 2/T})$, which are the cost of not knowing the true GP hyperparameters in advance. Note that in our theoretical analysis, we assume that the estimate obtained by the true loss function is consistent. This assumption is reasonable as it is the central analysis of most standard supervised learning methods. When we have an infinite number of i.i.d. observations, we could learn the hyperparameters accurately. It is worth emphasizing that with this assumption for the \textit{finite} observations, traditional regret bound analysis as in \citet{Srinivas_2010Gaussian} still does not hold, whilst our regret bound in Theorem \ref{theorem:final-convergence} holds.

\begin{figure*}[t]
    \centering
    \includegraphics[trim=0.1cm 0.4cm 0cm  0cm, clip, width=0.24\textwidth] {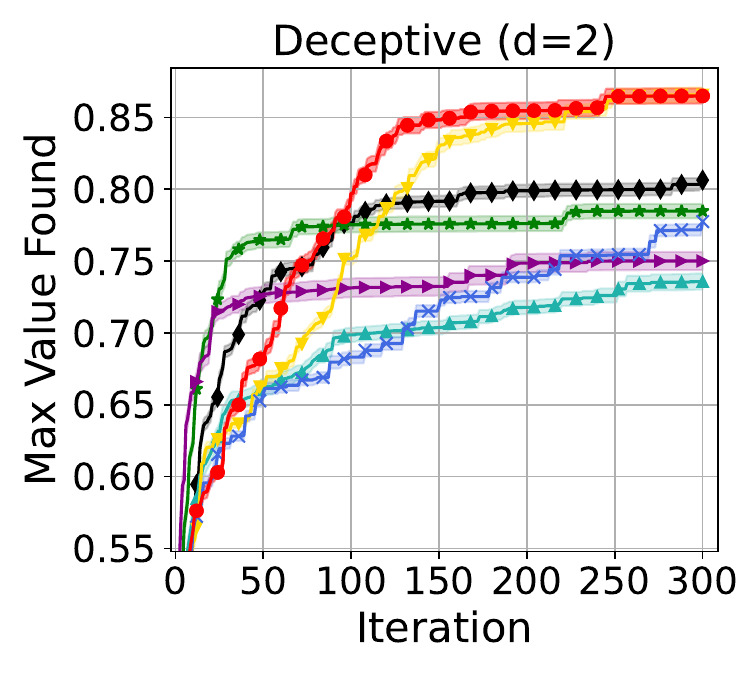}
        \includegraphics[trim=0.1cm 0.4cm 0cm  0cm, clip, width=0.24\textwidth]{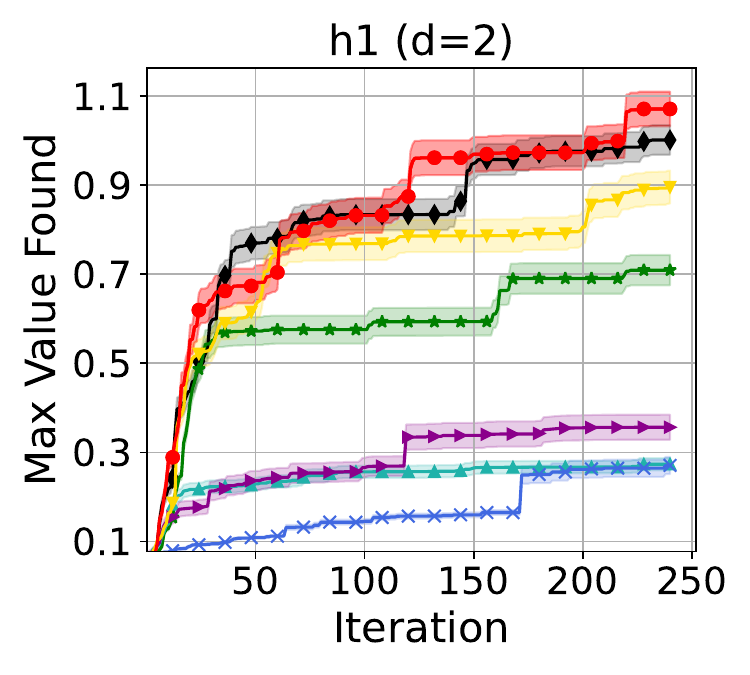}
        \includegraphics[trim=0.2cm 0.4cm 0cm  0cm, clip, width=0.24\textwidth]{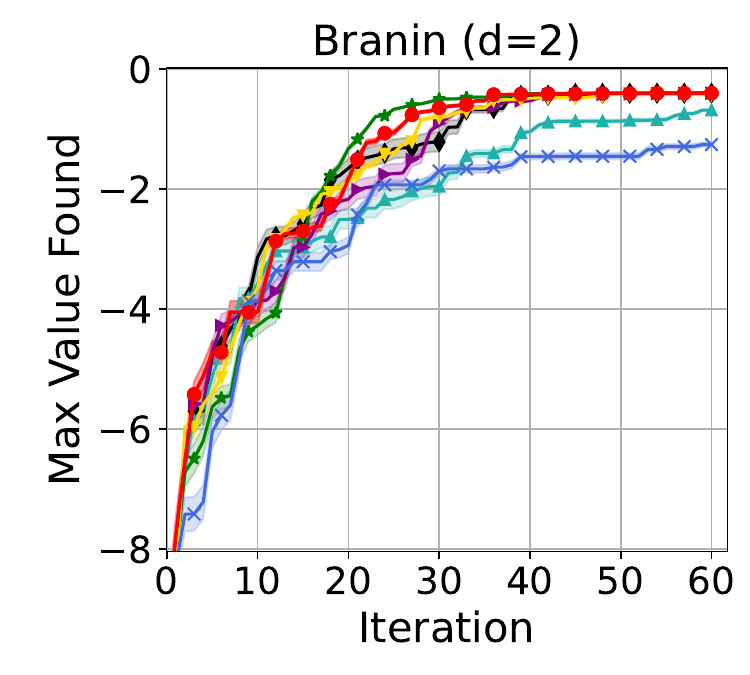}
    \includegraphics[trim=0.1cm 0.4cm 0cm  0cm, clip, width=0.24\textwidth]{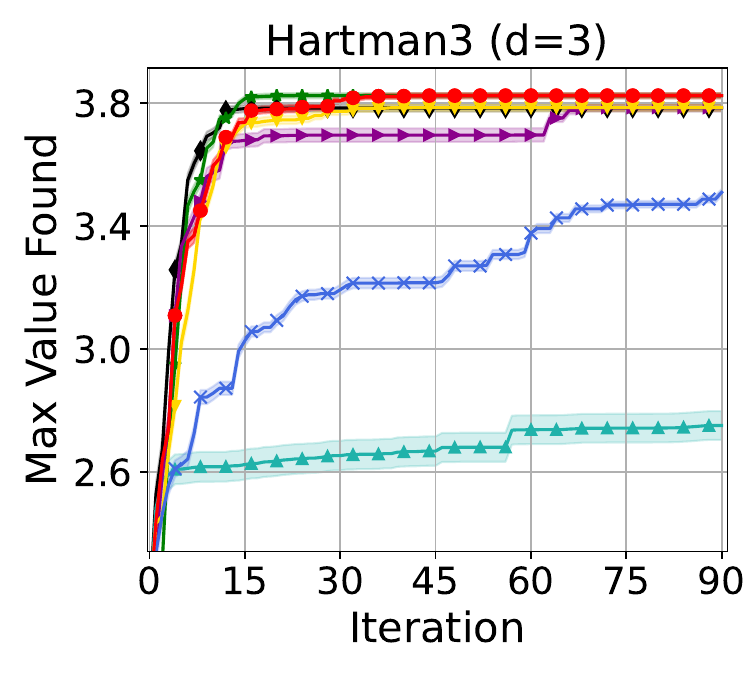}

    \includegraphics[trim=0cm 0.4cm 0cm  0cm, clip, width=0.32\textwidth]{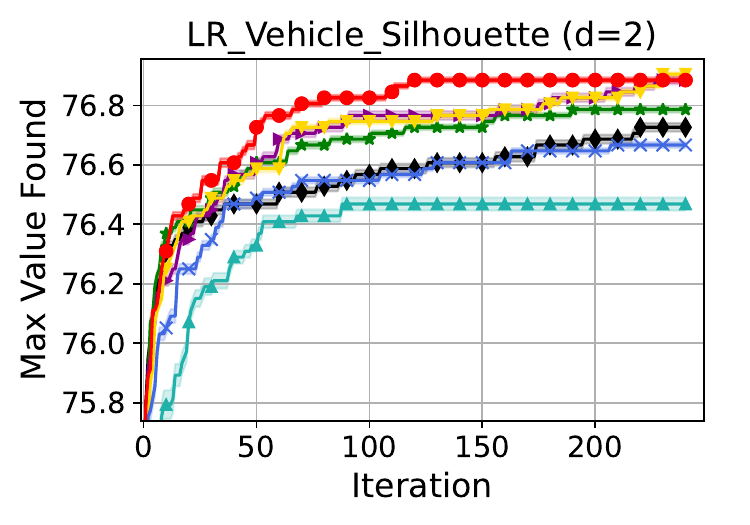}
        \includegraphics[trim=0cm 0.4cm 0cm  0cm, clip, width=0.32\textwidth]{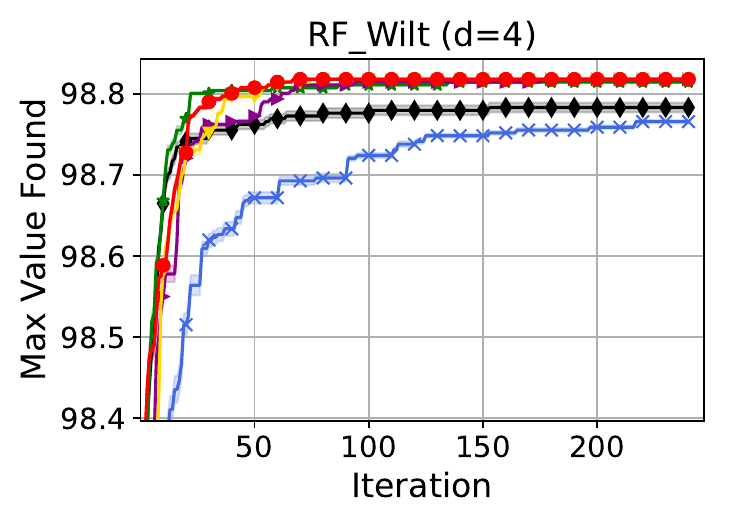}
    \includegraphics[trim=0cm 0.4cm 0cm  0cm, clip, width=0.32\textwidth]{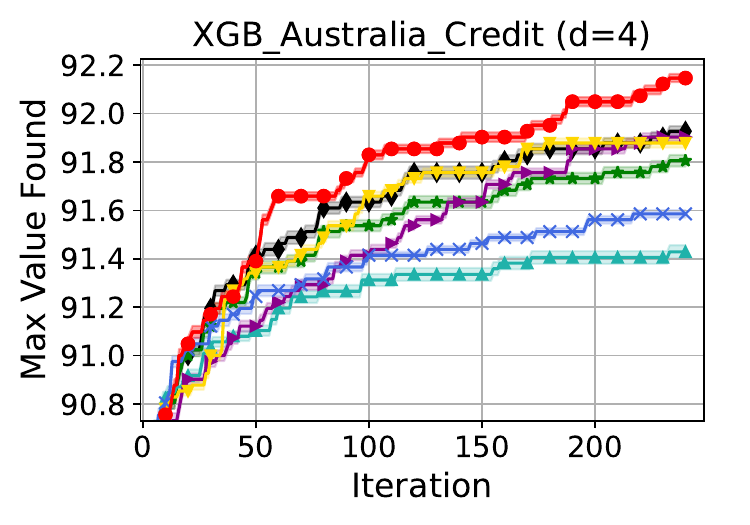}
    \includegraphics[trim=0cm 0cm 0cm  0cm, clip, width=0.95\linewidth]{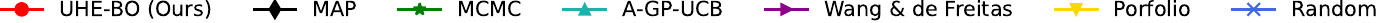} 

    \caption{Results on synthetic (\textbf{Top}) and real-world benchmarks (\textbf{Bottom}). Lines and shaded areas denote mean $\pm$ 1 standard error. Experiments are repeated 20 times.
    }    
    \label{fig:syn-realproblem}
   \vspace{-2mm}
\end{figure*}



\vspace{-0.1cm}
\section{Experiments} \label{sec:experiment}
\vspace{-0.2cm}

\textbf{Experimental Setup.}
All experimental results are averaged over $20$ independent runs with different random seeds.  We use Mat\'{e}rn 5/2 ARD kernels for the GP in all methods \cite{Rasmussen_2006gaussian}. For the MCMC approach, we generate $200$ burn-in and $10$ real MCMC samples. 



\textbf{Baselines.} We compare our proposed \acrshort{UHE} against six state-of-the-art methods for estimating hyperparameters for GP in BO: (1) \textbf{MAP}; (2) \textbf{MCMC} via HMC \citep{Neal2011HMC}; (3) \textbf{A-GP-UCB}: adaptive GP-UCB for BO with unknown hyperparameters \citep{Berkenkamp2019UnknownHyper}; (4) \textbf{Wang \& de Freitas' method}: BO with unknown hyperparameters \cite{Wang_2014Theoretical}; (5) \textbf{Portfolio}: described in \citet{Hoffman2011Portfolio}; (6) \textbf{Random}: the data points are chosen randomly. We are unable to compare with \citet{BogunovicK21} which also tackles the problem of BO with unknown GP hyperparameters. There were no empirical experiments conducted in \cite{BogunovicK21} so it is unclear how the methods can be implemented and empirically evaluated.


More details about the experiment setup and the benchmarks are in App. \ref{appendix_sec:exp_setup}.

\subsection{Synthetic and Real-world Benchmarks}

\paragraph{Synthetic Problems.}
In Fig. \ref{fig:syn-realproblem} (Top), our \acrshort{UHE} is significantly better than all methods on Deceptive and h1 whilst performing comparably on Branin and Hartman3. MAP and MCMC do not take into account the inconsistency and uncertainty issues of estimating GP hyperparameters, thus their performance is limited. These methods perform well on Branin and Hartman3, which is reasonable since these functions are known to be easy to optimize, however, they perform badly on Deceptive and h1. Other methods that are proposed to handle unknown GP hyperparameters are not as efficient as our method. \acrshort{UHE} outperforms A-GP-UCB and Wang \& de Freitas' most of the time. Although both techniques have sub-linear convergence guarantees, their empirical performance is conservative, as they tend to over-explore the search domain. This is due to their design of reducing the lengthscales monotonously, leading to the overestimation of the uncertainty. Finally, our \acrshort{UHE} also outperforms Porfolio and Random significantly.




\paragraph{Real-world Problems.}
In Fig. \ref{fig:syn-realproblem} (Bottom), we plot the results achieved by all methods. For real-world problems, the objective functions are often complex, our method \acrshort{UHE} still outperforms other approaches. For Logistic Regression with the Vehicle Silhouettes dataset and the XGB Classification with the Australia Credit dataset, \acrshort{UHE} outperforms existing methods significantly. For Random Forest with the Wilt dataset, \acrshort{UHE} is better than most of the approaches, except MCMC when it performs similarly. The Random method performs poorly in all cases. 

\subsection{Ablation Studies} \label{sec:ablation}

\begin{figure*}[t]
    \centering
    \includegraphics[trim=0cm 0cm 0cm  0cm, clip, width=0.99\linewidth]{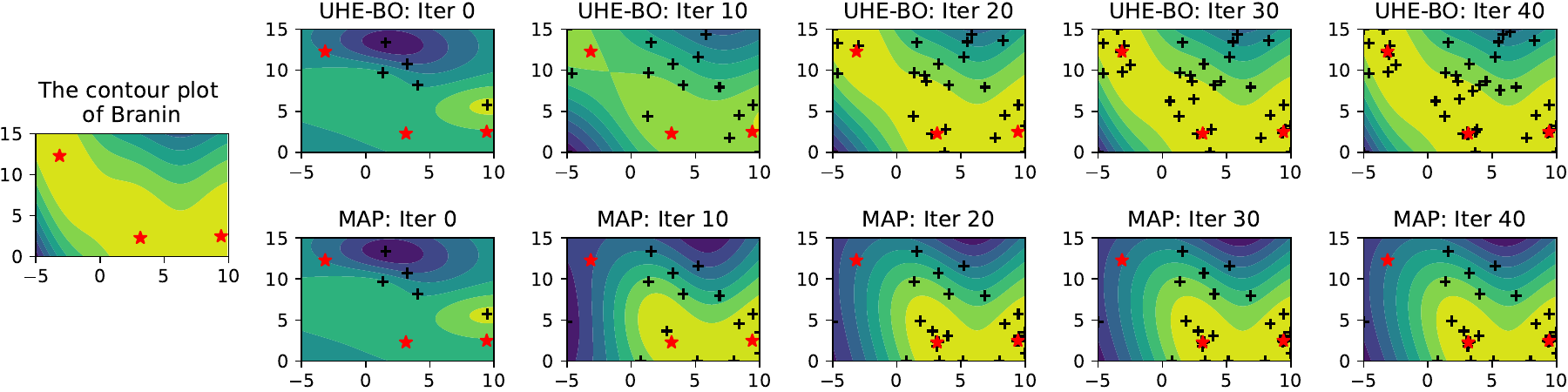}
    \caption{\textbf{Bottom}: Observed data (black crosses) and the GP mean by MAP. The GP hyperparameter estimation by MAP is incorrect due to a biased data collection process, resulting in the GP mean being different from the true function (in \textbf{Left}). \textbf{Top}: with our proposed method, consistent GP hyperparameter estimation can be obtained, and thus, the GP mean is more accurate.
    }     
    \label{fig:ablation-gpest}
\end{figure*}

\textbf{Consistency Correction in 2D.} In Fig. \ref{fig:ablation-gpest} \textbf{Top}, we plot the data sampled in each iteration and the corresponding GP estimation by our proposed method \acrshort{UHE}. In the \textbf{Bottom}, we show similar results using MAP estimation. It can be seen that our random sampling strategy and the consistency correction method help to improve the estimation of the GP hyperparameters and make the GP estimation more accurate, compared to the ground truth in \textbf{Left}. 

\begin{figure*}[t]
\vspace{-0.1cm}
    \centering
    \includegraphics[trim=0cm 0.4cm 0cm  0cm, clip, width=0.24\textwidth] {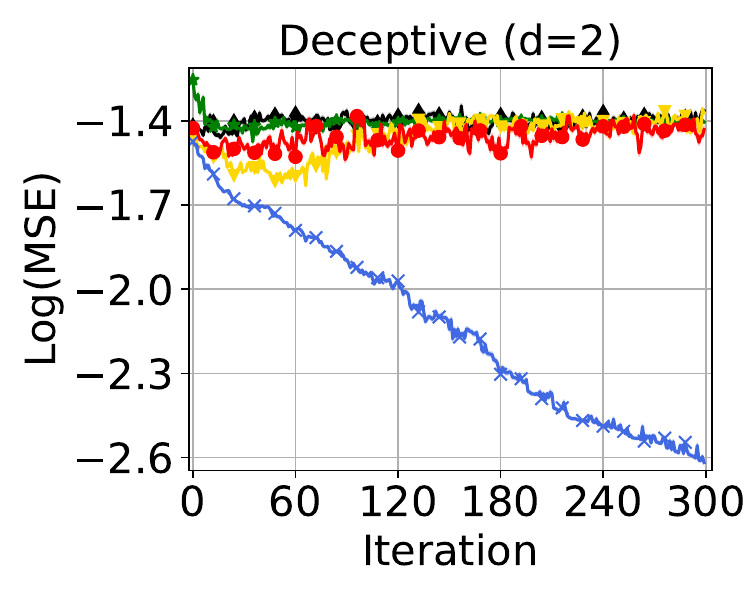}
        \includegraphics[trim=0cm 0.4cm 0cm  0cm, clip, width=0.24\textwidth]{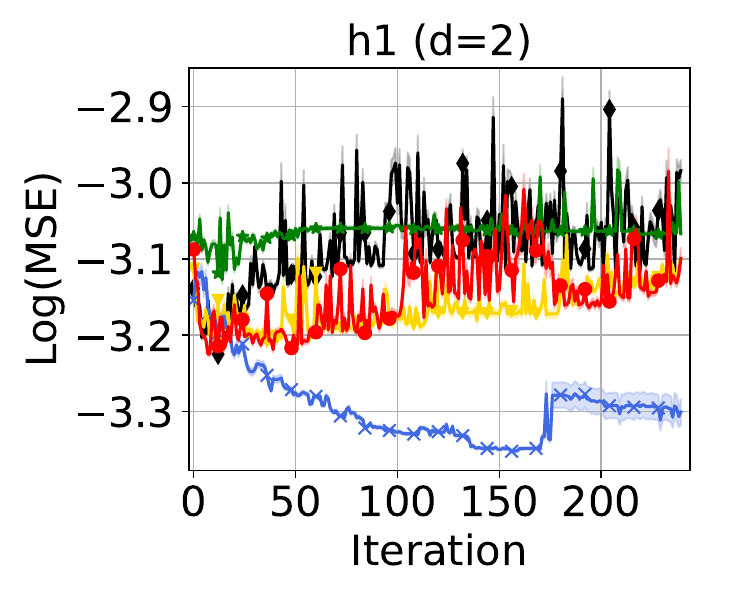}
        \includegraphics[trim=0cm 0.4cm 0cm  0cm, clip, width=0.24\textwidth]{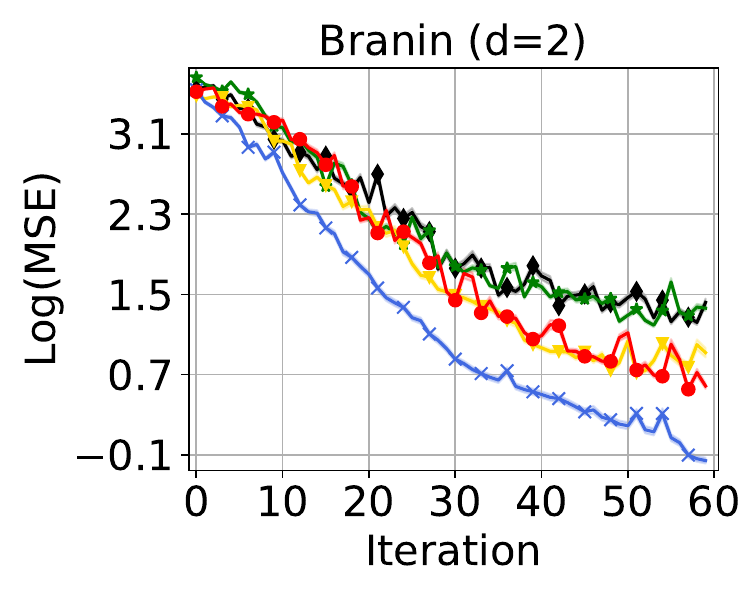}
    \includegraphics[trim=0cm 0.4cm 0cm  0cm, clip, width=0.24\textwidth]{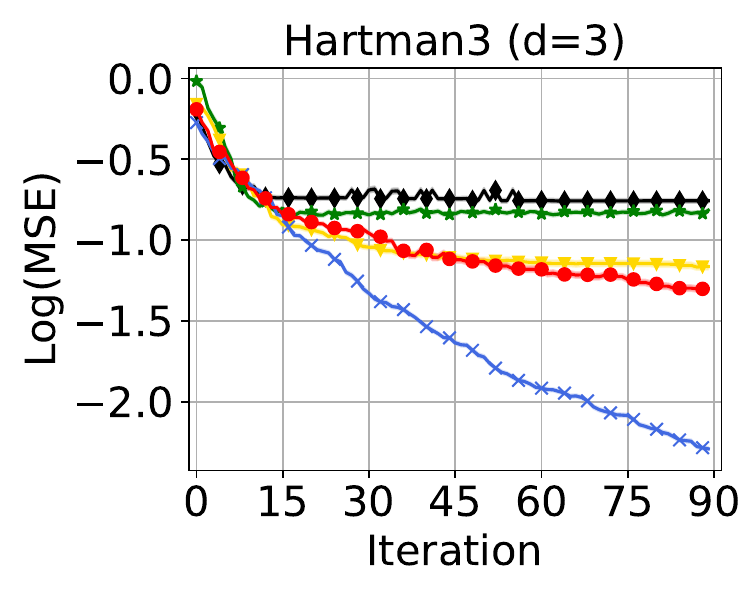}
    \vspace{-0.1cm}
    \includegraphics[trim=0cm 0cm 0cm  0cm, clip, width=0.7\linewidth]{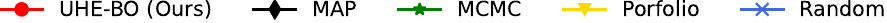} 
    
    \caption{MSE results on synthetic benchmarks. Lines and shaded areas are mean $\pm$ 1 standard error.
    }    
    \label{fig:mse-syntheticproblem}
   \vspace{-4mm}
\end{figure*}

\textbf{Evaluation of the GP Fitting.} In Fig. \ref{fig:mse-syntheticproblem}, for each synthetic problem, we generate a large number of data points in the search domain (approximately 216000 to 250000 data points), and compute and plot the mean square errors (MSEs) between the GP predictions and the objective function values. Random gives the most accurate GP predictions, which makes sense, as the GP estimation is more accurate when the data is i.i.d, however, as shown in Fig. \ref{fig:syn-realproblem}, Random performs very badly when finding the global optimum. Our \acrshort{UHE} estimates the GPs more accurately than all baseline methods (except Random) and notes that it also performs the best in finding the global optimum (see Fig. \ref{fig:syn-realproblem}). \textbf{Porfolio} also performs well but it's not as good as \acrshort{UHE}. Note we only plot for synthetic problems as it is not possible to evaluate a large number of data points for real-world problems.


\textbf{The Effect of Each Component in \acrshort{UHE}.} We investigate the effect of: (i) random sampling based on EXP3 (Sec. \ref{sec:samplingiid}) and (ii) using our proposed loss function  (Sec. \ref{sec:unbiasedloss}). In Fig. \ref{fig:ablation-component} (App. \ref{appendix_sec:add_exps}), we plot the performance of the method including both steps (\acrshort{UHE}), the variant only including (i) (Random+EXP3), the variant only including (ii) (\acrshort{RDEXP3}). Results show that each component contributes to the success of the proposed method. The performance of the method \acrshort{RDEXP3} is quite conservative as it samples data points randomly every $2$ iterations. In contrast, \acrshort{UHE} is more efficient because it learns to switch between decisions using EXP3. 

\textbf{Running Time Comparison.} In Fig. \ref{fig:ablation-component} (App. \ref{appendix_sec:add_exps}), we show the running time of our method versus two other standard methods: MAP and MCMC on three problems (Deceptive, Hartman3, and RF Wilt). The running time of our approach is similar to MAP whilst much faster than MCMC.

\vspace{-0.3cm}
\section{Conclusion and Limitation} 
\label{sec:conclusion}
\vspace{-0.25cm}

We have presented a new BO method for handling the scenarios when the true GP hyperparameters are unknown.
The main idea is to correct the bias issue of the data caused by the BO sampling process. To do so, we suggest sampling and adding i.i.d. data points to the BO process via a multi-armed bandit technique (EXP3). We then propose a new training loss function to achieve consistent estimations of the GP hyperparameters from the observed data. We further provide the theoretical analysis to upper bound the expected simple regret of our proposed method. Finally, we demonstrate the efficacy of our proposed method on a variety of synthetic and real-world problems.

\vspace{-0.25cm}
\paragraph{Limitation.}
Despite the great advantages, we identify several limitations. First, our theoretical analysis relies on the assumption that the estimate obtained by a hyperparameter loss function constructed from i.i.d. data is consistent. This can be true for some kernels and some types of hyperparameters~\cite{Bachoc2013MLE}, but not all kernels. Second, our sub-linear convergence rate is for the expected simple regret, and applies when the iterations $T$ are large values, so for a small evaluation budget, our proposed method might not achieve a sub-linear regret bound. Finally, our method is not effective for high-dimensional optimization problems as the random sampling strategy is generally over-exploring in high-dimensional search space. In our future work, we aim to address these limitations.



\bibliography{references}
\bibliographystyle{plainnat}

\newpage
\appendix

\section{Preliminaries} \label{sec:prelim}

\subsection{Gaussian Processes}

A Gaussian process (GP) defines a probability distribution over functions $f$ under the assumption that any finite subset $\lbrace (x_i, f(x_i) \rbrace$ follows a normal distribution \cite{Rasmussen_2006gaussian}. Formally, a GP is defined  as $f(x)\sim \mathcal{GP}\left(\mu\left(x\right),k\left(x,x'\right)\right)$, where $\mu(x)=\mathbb{E}\left[f\left(x\right)\right]$ is the mean function and $k(x,x')=\mathbb{E}\left[(f\left(x\right)-\mu\left(x\right))(f\left(x'\right)-\mu\left(x'\right))^{T}\right]$ is the covariance function \cite{Rasmussen_2006gaussian}.

We have the joint multivariate Gaussian distribution between the observed data $\boldsymbol{f} = [f(x_1),\dots,f(x_N)]$ $(\boldsymbol{f} \in \mathbb{R}^N)$ and a new data point $\bx_{\text{n}},f_{\text{n}}=f\left(\bx_{\text{n}}\right)$ -- assuming zero mean prior $m(x)=0$ for simplicity, 
\begin{align}
\left[\begin{array}{c}
\boldsymbol{f}\\
f_{\text{n}}
\end{array}\right] & \sim\mathcal{N}\left(0,\left[\begin{array}{cc}
\bK & \bk_{\text{n}}^{T}\\
\bk_{\text{n}} & k_{\text{n}}
\end{array}\right]\right),\label{eq:p(f|f*)}
\end{align}
 where $k_{\text{n}}=k\left(x_{\text{n}},x_{\text{n}}\right)$, $\bk_{\text{n}}=[k\left(x_{\text{n}},x_{i}\right)]_{\forall i\le N}$
and $\bK=\left[k\left(x_{i},x_{j}\right)\right]_{\forall i,j\le N}$. Combining Eq. (\ref{eq:p(f|f*)}) with the fact that $p\left(f_{\text{n}}\mid\boldsymbol{f}\right)$ follows a univariate Gaussian distribution $\mathcal{N}\left(\mu\left(x_{\text{n}}\right),\sigma^{2}\left(x_{\text{n}}\right)\right)$, the  posterior mean and variance can be computed as,
\begin{align*}
\mu\left(x_{\text{n}}\right)= & \mathbf{k}_{\text{n}} \left[ \mathbf{K} + \sigma^2 \idenmat_N \right]^{-1}\mathbf{y},\\
\sigma^{2}\left(x_{\text{n}}\right)= & k_{\text{n}}-\mathbf{k}_{\text{n}} \left[ \mathbf{K} + \sigma^2 \idenmat_N \right]^{-1} \mathbf{k}_{\text{n}}^{T},
\end{align*}
with $\by = [y_1,\dots,y_N]$.

As GPs give full uncertainty information with any prediction, they provide a flexible nonparametric prior for Bayesian optimization. We refer the interested readers to \cite{Rasmussen_2006gaussian} for further details on GP.

\subsection{GP Log Marginal Likelihood \& Log Posterior Distribution Function} \label{app_sec_GP_LogMarginalLLK}
Let us denote $\theta$ be the GP hyperparameter, and given the observed data $\{(x_i, y_i)\}_{i=1}^N$ then the log marginal likelihood of the GP hyperparameter $\theta$ is available in closed form \cite{Rasmussen_2006gaussian},
\begin{equation} \label{eq:loglikelihood}
\begin{aligned}
\mathcal{L}_{\textsc{MLL}}(\theta)&=\log p\left(\by \mid \bx,\theta\right) \\
&= -\frac{1}{2}\by^{T}\left(\bK_{\theta}+\sigma^{2}\idenmat_{N}\right)^{-1}\by \\
&\quad -\frac{1}{2}\log\left|\bK_{\theta}+\sigma^{2}\idenmat_{N}\right| - \frac{N}{2}\log (2\pi),
\end{aligned}
\end{equation}
 where $\sigma^2$ is the measurement noise variance, $\idenmat_{N}$ is the identity matrix, $\bK_{\theta} \in \mathbb{R}^{N\times N}$ is the covariance matrix w.r.t kernel $k_{\theta}(x,x')$, $\bx$ is the vector of input, $\by$ is the vector of observed output, and $N$ is the number of points in the observed data. With this formula, the Maximum Likelihood Estimate (MLE) can be computed as $\hat{\theta}_{\textsc{MLE}} = {\operatorname{argmax}}_{\theta} \ \mathcal{L}_{\textsc{MLL}}(\theta)$.
 
 If a prior distribution over $\theta$, $p(\theta)$, is available, then the log posterior distribution function can be computed as,
 \begin{equation} \label{eq:logpostdist}
\begin{aligned}
\mathcal{L}_{\textsc{MAP}}(\theta)&=\log p\left(\by \mid \bx,\theta\right) + \log p(\theta) \\
&= -\frac{1}{2}\by^{T}\left(\bK_{\theta}+\sigma^{2}\idenmat_{N}\right)^{-1}\by \\
& \quad -\frac{1}{2}\log\left|\bK_{\theta}+\sigma^{2}\idenmat_{N}\right| - \frac{N}{2}\log (2\pi) + \log p(\theta).
\end{aligned}
\end{equation}
With this log posterior distribution function, the MAP estimate can be computed as $\hat{\theta}_{\textsc{MAP}} = {\operatorname{argmax}}_{\theta} \ \mathcal{L}_{\textsc{MAP}}(\theta)$.

\subsection{Bayesian Optimization} \label{sec:app-bo}
Bayesian optimization solves the optimization problem in Eq. (\ref{eq:bo}) in a sequential manner. First, at iteration $t$, a surrogate model is used to approximate the objective function $f(.)$ over the current observed data $D_{t-1} = \{x_i, y_i\}_{i=1}^{\vert D_{t-1} \vert}$ where $y_i=f(x_i) + \epsilon_i$, and $\epsilon_i \sim \mathcal{N}(0, \sigma^2)$. An acquisition function is constructed using the surrogate model to select the next data point $x_t$ to sample. The objective function $f(\cdot)$ is evaluated at $x_t$, and the pair $\{x_t, y_t\}$ is added to the observation set, i.e., $D_{t} = D_{t-1} \cup \{x_t, y_t\}$. These steps are repeated until the evaluation budget is exhausted, at which the data point with the highest function value is selected.
 
\subsection{Maximum Information Gain}
For a GP with hyperparameter $\theta$, the maximum mutual information $\gamma_T(\theta)$ is defined as follows \citep{Srinivas_2010Gaussian},
\begin{equation} \label{eq-mig}
\gamma_T(\theta) := \max_{A \subset \mathcal{X}, \vert A \vert =T} \dfrac{1}{2} \log \det (\idenmat_T + \sigma^{-2} \bK_{\theta}),
\end{equation}
where $\bK_{\theta} = \lbrack k_{\theta}(x,x') \rbrack_{x,x' \in A}$ and $\idenmat_T$ is the identity matrix of size $T \times T$.

\subsection{The EXP3 Algorithm} \label{sec:app-exp3}
EXP3, which stands for Exponential-weight algorithm for Exploration and Exploitation, is an algorithm that tackles the multi-armed bandit (MAB) problem \cite{Auer2002nonstochastic}. An MAB problem is characterized by $M$ possible arms and an assignment of rewards, i.e., an infinite sequence $\mathbf{r}_1, \mathbf{r}_2,\dots$ of vectors $\mathbf{r}_t=(r^1_t, r^2_t,\dots,r^M_t)$ with $r^m_t$ being denoted as the reward obtained if the arm $m \in \{1,\dots,M\}$ is chosen at time step $t$. Assume the player knows the number of $M$ arms, and also assume that after each time step $t$, the player only knows the rewards $r^{a_1}_1,\dots,r^{a_t}_t$ of the previous chosen actions $a_1,\dots,a_t$. The goal of an MAB algorithm is to identify the sequence of arms so as to maximize the total reward $\sum_{t=1}^T r^{a_t}_t$ for any $T>0$.

The EXP3 method is an effective and efficient algorithm to solve the MAB problem. The algorithm is described in Algorithm \ref{alg:exp3} \cite{Auer2002nonstochastic}. An advantage of the EXP3 algorithm is that it can be applied to the partial information setting, that is, it only requires the reward of the chosen action in each iteration. Besides, it has been shown to have strong convergence property. In particular, let us denote $R_{\text{EXP3}}(T)=\sum_{t=1}^T r^{a_t}_t$ as the total reward at time horizon $T$ of EXP3. Let us denote $R_{\text{max}}(T)=\max_{m} \sum_{t=1}^T r^m_t$ as the total reward of the single globally best action at time horizon $T$. Then we have that, for any $M>0$ and for any $\gamma \in (0,1]$, $R_{\max}(T)-\mathbb{E}[R_{\text{EXP3}}(T)] \leq (e-1)\gamma R_{\max} + \dfrac{M \ln M}{\gamma}$ holds for any assignment of rewards and for any $T > 0$ (Theorem 3.1 \cite{Auer2002nonstochastic}). Here the expectation in the inequality is due to the randomness of EXP3 when selecting the arm in each iteration.

A simpler bound of this bound can be derived as follows, for any $T>0$, assume that $g$ is a value that is smaller than $R_{\max}$ and that algorithm EXP3 is run with the input parameter $\gamma = \min \{1, \sqrt{M\ln M/ \bigl((e-1)g \bigr)}\}$, then 
\begin{equation}
    \begin{aligned}
    R_{\max}-\mathbb{E}[R_{\text{EXP3}}(T)] & \leq 2\sqrt{e-1}\sqrt{gM \ln M} \\
    & \leq 2.63 \sqrt{gM\ln M}
    \end{aligned}
\end{equation}
holds for any assignment of rewards (Corollary 3.2 \cite{Auer2002nonstochastic}). Finally, it's worth noting that in the standard EXP3 algorithm, the rewards $r^m_t$ are in the range $[0,1]$. For the rewards instead in the arbitrary range $[a,b]$, then the same bound with the rhs terms multiplying with a constant factor can be shown. 

\begin{algorithm}[tb]
   \caption{EXP3 Algorithm \cite{Auer2002nonstochastic}}
   \label{alg:exp3}
\begin{algorithmic}[1]
   \STATE {\bfseries Input:} The real value $\gamma \in (0,1]$
   \STATE {\bfseries Output:} The arm at each iteration so as to maximize the total reward
   \STATE Initialize $\omega_0^m=1$ for $m=1,\dots,M$
   \FOR {$t = 1, 2, \dots $}
   \STATE Set $p^i_t = \dfrac{(1-\gamma)\omega^m_{t-1}}{\sum_{j=1}^K \omega^m_{t-1}} + \gamma/M, \quad \forall m=1,\dots,M$
   \STATE Draw an arm $a_t \in \{1,\dots,M\} \sim [p^1_t,\dots,p^M_t]$
   \STATE Receive the reward $r_t^{a_t} \in [0,1]$ of the chosen action $a_t$
   \STATE For $m=1,\dots,M$, set
   \begin{equation} \nonumber
   \begin{aligned}
       \hat{r}^m_t &= \begin{cases}
                        r^m_t/p^m_t \quad \text{if}\quad m=a_t \\
                        \quad \quad    0 \quad \quad \quad \ \text{otherwise}, 
                        \end{cases} \\
       \omega^m_t &= \omega^m_{t-1} \text{exp} \bigl(\gamma \hat{r}^m_t/M \bigr)
   \end{aligned}
   \end{equation}
   \ENDFOR
\end{algorithmic}
\end{algorithm}

\section{Theoretical Derivations} \label{appendix_sec:theory}

\subsection{Proof of Proposition \ref{prop:newloss}} \label{app:prop64-proof}


\subsubsection{Proving Bullet Point 1 in Proposition \ref{prop:newloss}} \label{sec:proof-propnewloss-1}

\begin{proof}

In this proof, we use Assumptions \ref{assum:l-loss} and \ref{assum:f-smoothness}.

 For a value $\epsilon$, let us denote $T_{\epsilon}$ as the value such that when we divide the search space $\mathcal{X}$ into $T_{\epsilon}$ equal parts, the maximum distance between any data points in each part is equal or less than $\epsilon$. Note that the observed dataset $D_{t-1}$ constructed by \acrshort{RDEXP3} consists of $\lfloor (t-1)/2 \rfloor$ random data points. Therefore, the probability of a part in $T_{\epsilon}$ parts to not have any random data points from $D_{t-1}$ is $(1-1/T_{\epsilon}) ^{\lfloor (t-1)/2 \rfloor}$. Then the probability of existing one of the $T_{\epsilon}$ parts to not have any random data points from $D_{t-1}$ is $T_{\epsilon}(1-1/T_{\epsilon}) ^{\lfloor (t-1)/2 \rfloor}$. Thus, the probability that all $T_{\epsilon}$ parts to have at least one random data point from $D_{t-1}$ is $1 - T_{\epsilon}(1-1/T_{\epsilon}) ^{\lfloor (t-1)/2 \rfloor}$. Similarly, the probability that all parts in $T_{\epsilon}$ parts to have at least one data point from the random dataset $\{x^\prime_j$\}$_{j=1}^{M_t}$ is $1 - T_{\epsilon}(1-1/T_{\epsilon}) ^{M_t}$.
 
 Finally, the probability $\forall x^\prime_j \in$ $\{x^\prime_j$\}$_{j=1}^{M_t}$ to be in the same part with at least one data point from $D_{t-1}$ is $p_{\epsilon}^t = (1 - T_{\epsilon}(1-1/T_{\epsilon}) ^{\lfloor (t-1)/2 \rfloor}) \times (1 - T_{\epsilon}(1-1/T_{\epsilon}) ^{M_t})$. Since the maximum distance between any two data points in each part is $\le \epsilon$, then $\forall x^\prime_j \in$ $\{x^\prime_j$\}$_{j=1}^{M_t}$, with probability (w.p.) $p_{\epsilon}^t$, $\exists \hat{x}^\prime_j \in D_{t-1}$: $|| x^\prime_j - \hat{x}^\prime_j ||_1 \le \epsilon$.
 
 Now, if we choose $\epsilon= \mathcal{O}(t^{-1/2d})$, then there exists the choice of $\epsilon$ such that $T_{\epsilon} = (t^{-1/2d})^{-d} = t^{1/2}$. With this choice, the probability $p_{\epsilon}^t$ becomes $(1 - t^{1/2}(1-1/t^{1/2}) ^{\lfloor (t-1)/2 \rfloor})(1 - t^{1/2}(1-1/t^{1/2}) ^{M_t})$. Note that it's easy to prove this probability converges to 1 when $t$ goes to infinity. We do a quick proof here. Let us consider $t^{1/2}(1-1/t^{1/2}) ^{\lfloor (t-1)/2 \rfloor}$, we will prove this term converges to 0 when $t$ goes to infinity. Taking log of this term results in $1/2\log(t) + \lfloor (t-1)/2 \rfloor\log(1-1/t^{1/2})$. Using L'Hôpital's rule, we have that the second term converges to $-1/4 t^{1/2}$ when $t$ goes to infinity. Therefore, combining this with $1/2\log(t)$ results in the log of $t^{1/2}(1-1/t^{1/2}) ^{\lfloor (t-1)/2 \rfloor}$ converges to $-\infty$, and thus, the probability $t^{1/2}(1-1/t^{1/2}) ^{\lfloor (t-1)/2 \rfloor}$ converges to 0, and thus the probability $1-t^{1/2}(1-1/t^{1/2}) ^{\lfloor (t-1)/2 \rfloor}$ converges to 1. The probability $(1 - t^{1/2}(1-1/t^{1/2}) ^{M_t})$ can be derived in a similar manner since $M_t \geq \vert D_t \vert$.

In summary, with probability $(1 - t^{1/2}(1-1/t^{1/2}) ^{\lfloor (t-1)/2 \rfloor})(1 - t^{1/2}(1-1/t^{1/2}) ^{M_t})$, we have that $\forall x^\prime_j \in$ $\{x^\prime_j$\}$_{j=1}^{M_t}$, $\exists \hat{x}^\prime_j \in D_{t-1}$: $|| x^\prime_j - \hat{x}^\prime_j ||_1 \le \mathcal{O}(t^{-1/2d})$. And note that the probability $(1 - t^{1/2}(1-1/t^{1/2}) ^{\lfloor (t-1)/2 \rfloor})(1 - t^{1/2}(1-1/t^{1/2}) ^{M_t})$ converges to 1 when $t$ goes to infinity.

Now let us consider the loss function $\mathcal{L}$, using Assumption \ref{assum:l-loss}, we have that, with probability $\ge 1-pe^{-(C_l/q)^2}$,

\begin{equation} \label{eq:loss-f}
    \begin{aligned}
        &\vert \mathcal{L} (\hat{\mathbf{y}}^\prime_t, \mathbf{x}^\prime_t, \theta) - \mathcal{L} (\mathbf{y}^\prime_t, \mathbf{x}^\prime_t, \theta) \vert \\
        & = \vert \frac{1}{M_t} \sum_{j=1}^{M_t} l(y(\hat{x}^\prime_j), x^\prime_j, \theta) - \frac{1}{M_t} \sum_{j=1}^{M_t} l(y(x^\prime_j), x^\prime_j, \theta) \vert \\
        & \leq \frac{1}{M_t} \sum_{j=1}^{M_t} \Big\vert l(y(\hat{x}^\prime_j), x^\prime_j, \theta) - l(y(x^\prime_j), x^\prime_j, \theta) \Big\vert \\
        & \leq \frac{C_l}{M_t} \sum_{j=1}^{M_t} \Big\vert f(\hat{x}^\prime_j) - f(x^\prime_j) \Big\vert.
    \end{aligned}
\end{equation}

Using Assumption \ref{assum:f-smoothness}, we have that, with probability greater than $1-dae^{-(L/b)^2}$, $\forall \hat{x}^\prime_j, x^\prime_j$, $\Big\vert f(\hat{x}^\prime_j) - f(x^\prime_j) \Big\vert \leq L \Vert \hat{x}^\prime_j - x^\prime_j \Vert_1$. Using the Fr\'{e}chet inequality and combining with the inequality in Eq. (\ref{eq:loss-f}), we have that, with probability greater than $1 - dae^{-(L/b)^2} - pe^{-(C_l/q)^2}$, $
\vert  \mathcal{L} (\hat{\mathbf{y}}^\prime_t, \mathbf{x}^\prime_t, \theta) - \mathcal{L} (\mathbf{y}^\prime_t, \mathbf{x}^\prime_t, \theta) \vert \leq \frac{L C_l}{M_t} \sum_{j=1}^{M_t} \Vert \hat{x}^\prime_j - x^\prime_j \Vert_1.
$

Finally, combining with the derivation regarding the distance between any $\hat{x}^\prime_j$ and $x^\prime_j$, with probability $(1 - dae^{-(L/b)^2} - pe^{-(C_l/q)^2})(1 - t^{1/2}(1-1/t^{1/2}) ^{\lfloor (t-1)/2 \rfloor})(1 - t^{1/2}(1-1/t^{1/2}) ^{M_t})$, we have that,
$
    \vert  \mathcal{L} (\hat{\mathbf{y}}^\prime_t, \mathbf{x}^\prime_t, \theta) - \mathcal{L} (\mathbf{y}^\prime_t, \mathbf{x}^\prime_t, \theta) \vert \leq \mathcal{O}(L C_l t^{-1/2d}).
$

Note in the above, the two events: (1) random data points come closer to each other and (2) the behavior of the loss function and objective functions are independent (as the 1st event is random) so we can multiply the two probabilities. Furthermore, the probability $(1 - dae^{-(L/b)^2} - pe^{-(C_l/q)^2})(1 - t^{1/2}(1-1/t^{1/2}) ^{\lfloor (t-1)/2 \rfloor})(1 - t^{1/2}(1-1/t^{1/2}) ^{M_t})$ is a high probability when $t$ goes to infinity.

\end{proof}

\subsubsection{Proving Bullet Point 2 in Proposition \ref{prop:newloss}} \label{sec:app-lest}

\begin{proof}

Using a similar proof technique, if we also want a bound on the derivative $\partial \mathcal{L}/\partial \theta$ (in addition to the bound on the loss function $\mathcal{L}$), then we have that, with probability $(1 - dae^{-(L/b)^2} - pe^{-(C_l/q)^2} - p^\prime e^{-(C_{l^\prime}/q^\prime)^2})(1 - t^{1/2}(1-1/t^{1/2}) ^{\lfloor (t-1)/2 \rfloor})(1 - t^{1/2}(1-1/t^{1/2}) ^{M_t})$, we have both the followings,
\begin{equation} \label{eq:loss-2-bound}
    \begin{aligned}
    \vert  \mathcal{L} (\hat{\mathbf{y}}^\prime_t, \mathbf{x}^\prime_t, \theta) - \mathcal{L} (\mathbf{y}^\prime_t, \mathbf{x}^\prime_t, \theta) &\vert \leq \mathcal{O}(L C_l t^{-1/2d}), \\
    \vert \partial \mathcal{L} (\hat{\mathbf{y}}^\prime_t, \mathbf{x}^\prime_t, \theta)/\partial \theta - \partial \mathcal{L} (\mathbf{y}^\prime_t, \mathbf{x}^\prime_t, \theta)/\partial \theta \vert &\leq \mathcal{O}(L C_l t^{-1/2d}).
    \end{aligned}
\end{equation}

With this, we have that there exists a value $T_1$ that $\forall t > T_1$, with the probability of $(1 - dae^{-(L/b)^2} - pe^{-(C_l/q)^2} - p^\prime e^{-(C_{l^\prime}/q^\prime)^2}) \prod_{i=T_1}^t(1 - i^{1/2}(1-1/i^{1/2}) ^{\lfloor (i-1)/2 \rfloor}) \prod_{i=T_1}^t (1 - i^{1/2}(1-1/i^{1/2}) ^{M_i})$, the inequalities in Eq. (\ref{eq:loss-2-bound}) are satisfied with all $t>T_1$.

Finally, using Theorem 7.17 in \cite{rudin1964principles}, we have that, with the probability of $(1 - dae^{-(L/b)^2} - pe^{-(C_l/q)^2} - p^\prime e^{-(C_{l^\prime}/q^\prime)^2}) \prod_{i=T_1}^t(1 - i^{1/2}(1-1/i^{1/2}) ^{\lfloor (i-1)/2 \rfloor}) \prod_{i=T_1}^t (1 - i^{1/2}(1-1/i^{1/2}) ^{M_i})$, the estimate obtained by minimizing the proposed loss function, $\hat{\theta}_t^{RA-BO} \rightarrow \theta^*$ for all $t>T_1$.

It's worth noting that based on the property that the tail of convergent series tends to 0, we have that $\prod_{i=T_1}^t(1 - i^{1/2}(1-1/i^{1/2}) ^{\lfloor (i-1)/2 \rfloor}) \prod_{i=T_1}^t (1 - i^{1/2}(1-1/i^{1/2}) ^{M_i}$ is close to $1$ when $t$ and $T_1$ are high. Therefore, the probability $(1 - dae^{-(L/b)^2} - pe^{-(C_l/q)^2} - p^\prime e^{-(C_{l^\prime}/q^\prime)^2}) \prod_{i=T_1}^t(1 - i^{1/2}(1-1/i^{1/2}) ^{\lfloor (i-1)/2 \rfloor}) \prod_{i=T_1}^t (1 - i^{1/2}(1-1/i^{1/2}) ^{M_i})$ is close to 1 when $t$ is high and the values $T_1, L, C_l, C_l^\prime$ are chosen to be large.

\end{proof}


\subsection{Proof of Theorem \ref{theorem:thetaconvergence}}

\begin{proof}


This is a result of the Bernstein-von Mises theorem \cite{Borwanker1971BvMTheorem, Vaart1998AsymStats}. To use this theorem, we need to have the following conditions,
\begin{itemize}
    \item The GP prior mean function is $0$;
    \item The prior distribution $p(\theta)$ of the GP hyperparameter $\theta$ is continuous and positive in an open neighborhood of the true parameter $\theta^*$;
    \item The loss function $\mathcal{L}(\hat{\by}_t, \bx'_t, \theta)$ satisfies that $\partial \mathcal{L}(\hat{\by}_t, \bx'_t, \theta) / \partial \theta$ and $\partial^2 \mathcal{L}(\hat{\by}_t, \bx'_t, \theta) / \partial \theta^2$ exist and are continuous in $\theta$;
    \item The Fisher information matrix $\hat{I}_t(\theta) = -\mathbb{E}_{\theta} [\partial^2 \mathcal{L}(\hat{\by}_t, \bx'_t, \theta) / \partial \theta^2]$ is continuous, positive, and upper bounded;
    \item The estimate $\hat{\theta}_t^{\acrshort{RDEXP3}} = {\operatorname{argmin}}_{\theta} \ \mathcal{L}(\hat{\by}_t, \bx'_t, \theta)$ is consistent.
\end{itemize}

Following Assumption \ref{assum:Bernstein-von-Mises}, the first two bullets are directly satisfied.

The condition in the third bullet automatically follows by the requirement in Assumption \ref{assum:Bernstein-von-Mises}, which states that the GP hyperparameter loss function $\mathcal{L}$ satisfies that $\partial \mathcal{L} \bigl(\by_t', \bx_t', \theta) / \partial \theta$ and $\partial^2 \mathcal{L} \bigl(\by_t', \bx_t', \theta) / \partial \theta^2$ exist and are continuous in $\theta$ for all $t$.  For the proposed loss function $\mathcal{L} \bigl(\hat{\by}_t', \bx_t', \theta)$, since we only substitute $\by'_t$ by $\hat{\by}_t'$, thus, the same properties can be concluded for $\mathcal{L} \bigl(\hat{\by}_t', \bx_t', \theta)$.

The requirement in the fifth bullet is satisfied based on the result of Proposition \ref{prop:newloss}. Specifically, the estimate obtained by \acrshort{RDEXP3} shares the same convergence property with the estimate obtained by the real loss function with probability $(1 - dae^{-(L/b)^2} - pe^{-(C_l/q)^2} - p^\prime e^{-(C_{l^\prime}/q^\prime)^2}) \prod_{i=T_1}^t(1 - i^{1/2}(1-1/i^{1/2}) ^{\lfloor (i-1)/2 \rfloor}) \prod_{i=T_1}^t (1 - i^{1/2}(1-1/i^{1/2}) ^{M_i})$.

In the below, we derive the proof for the fourth bullet, which states that the Fisher information matrix $\hat{I}_t(\theta)$ needs to be continuous, positive, and upper bounded. Using Assumption \ref{assum:f-smoothness}, we have that the true Fisher information matrix $I(\theta)$ is continuous, positive, and upper bounded. Using the same argument as in the proof of Proposition \ref{prop:newloss} (Sec. \ref{app:prop64-proof}), with probability $(1 - dae^{-(L/b)^2} - p^{\prime\prime} e^{-(C_{l^{\prime\prime}}/q^{\prime\prime})^2})(1 - t^{1/2}(1-1/t^{1/2}) ^{\lfloor (t-1)/2 \rfloor})(1 - t^{1/2}(1-1/t^{1/2}) ^{M_t})$, we have that,
$ \vert \partial^2 \mathcal{L} (\hat{\mathbf{y}}^\prime_t, \mathbf{x}^\prime_t, \theta)/\partial \theta^2 - \partial^2 \mathcal{L} (\mathbf{y}^\prime_t, \mathbf{x}^\prime_t, \theta)/\partial \theta^2 \vert \leq \mathcal{O}(t^{-1/2d}).$ 
Therefore, with large $t$, the Fisher information matrix $\hat{I}_t(\theta)$ is continuous, positive, and upper bounded. 

Combining the probabilities for fourth and fifth bullets, we have that, there exists $T_1$, with probability greater than $(1 - dae^{-(L/b)^2} - pe^{-(C_l/q)^2} - p^\prime e^{-(C_{l^\prime}/q^\prime)^2}) \prod_{i=T_1}^t(1 - i^{1/2}(1-1/i^{1/2}) ^{\lfloor (i-1)/2 \rfloor}) \prod_{i=T_1}^t (1 - i^{1/2}(1-1/i^{1/2}) ^{M_i})$, the conditions of the Bernstein-von Mises theorem are satisfied, therefore, with $t>T_1$, with probability greater than $(1 - dae^{-(L/b)^2} - pe^{-(C_l/q)^2} - p^\prime e^{-(C_{l^\prime}/q^\prime)^2}) \prod_{i=T_1}^t(1 - i^{1/2}(1-1/i^{1/2}) ^{\lfloor (i-1)/2 \rfloor}) \prod_{i=T_1}^t (1 - i^{1/2}(1-1/i^{1/2}) ^{M_i})$, we have $\hat{\theta}_t^{RA-BO} = \theta^* + \mathcal{O}(t^{-1/2})$.

Here we have the probability $(1 - dae^{-(L/b)^2} - pe^{-(C_l/q)^2} - p^\prime e^{-(C_{l^\prime}/q^\prime)^2} - p^{\prime\prime} e^{-(C_{l^{\prime\prime}}/q^{\prime\prime})^2})$ as we need to ensure that all the assumptions regarding the  Lipschitz continuity of $f, l, \partial l/\partial \theta, \partial^2 l/\partial \theta^2$ are satisfied. The probability $\prod_{i=T_1}^t(1 - i^{1/2}(1-1/i^{1/2}) ^{\lfloor (i-1)/2 \rfloor}) \prod_{i=T_1}^t (1 - i^{1/2}(1-1/i^{1/2}) ^{M_i})$ is the probability of two random data points come closer for large $t$ $(t>T_1)$ and large $T_1$. Similar to the discussion in the previous section, this probability is also high probability for large $t$ as we can choose the values $C_l, C_{l^\prime}, C_{l^{\prime\prime}}$ to be large to ensure $dae^{-(L/b)^2} + pe^{-(C_l/q)^2} + p^\prime e^{-(C_{l^\prime}/q^\prime)^2} + p^{\prime\prime} e^{-(C_{l^{\prime\prime}}/q^{\prime\prime})^2}$ to be small, and thus $(1 - dae^{-(L/b)^2} - pe^{-(C_l/q)^2} - p^\prime e^{-(C_{l^\prime}/q^\prime)^2} - p^{\prime\prime} e^{-(C_{l^{\prime\prime}}/q^{\prime\prime})^2})$  is close to $1$.


\end{proof}

\subsection{Proof of Theorem \ref{theorem:regret-random}}

\subsubsection{Proving Bullet Point 1 of Theorem \ref{theorem:regret-random}} \label{sec:app-lipschitzt}

\begin{proof}

Based on Theorem \ref{theorem:thetaconvergence}, we have that, there exists $T_1$ that $\forall t > T_1$, with probability greater than $(1 - dae^{-(L/b)^2} - pe^{-(C_l/q)^2} - p^\prime e^{-(C_{l^\prime}/q^\prime)^2}) \prod_{i=T_1}^t(1 - i^{1/2}(1-1/i^{1/2}) ^{\lfloor (i-1)/2 \rfloor}) \prod_{i=T_1}^t (1 - i^{1/2}(1-1/i^{1/2}) ^{M_i})$, $\hat{\theta}_t^{\text{\acrshort{RDEXP3}}} = \theta^* + \alpha_t t^{-1/2}$ with $\alpha_t \in \mathbb{R}^d$, and $\alpha_t$ is upper bounded by a constant vector.

Following Assumption \ref{assum:f-smoothness}, we have that the true hyperparameter $\theta^*$ satisfies the condition that its corresponding kernel $k_{\theta^*}(x,x')$ satisfies the following high probability bound on the derivatives of GP sample paths $f$: for some constants $a,b>0$, then $\text{Pr}\{ \sup_{\bx\in D} \vert \partial f/\partial x_h \vert > L\} \leq ae^{-(L/b)^2}, h=1,\dots,d$.

Note we have the property that the derivative functions $\partial / \partial x_h f(x)$ of a sample path $f$ drawn from a kernel $k(x,x^\prime)$ are sample paths drawn from the derivative GP with the derivative kernel $k_{\partial}(x,x^\prime)= \frac{ \partial^2 k (x,x^\prime)} { \partial x_h \partial x_h\prime }$ [36]. And note that the maximum of the $\partial / \partial x_h f(x)$ in the search domain are the Lipschitz constants of the sample path $f(x)$ [36]. For kernels that are stationary and twice differentiable in $\theta$, using Bochner's theorem, the spectral density of the kernel is also continuous in $\theta$. Since a sample path is a linear combination of the spectral density, therefore, if we have Assumption \ref{assum:f-smoothness} satisfied, we will also then have that, there exists $T_2 (T_2 \geq T_1)$ that $\forall t>T_2$, the estimate $\hat\theta_t^{RA-BO}$ will satisfy the requirements that for all sample paths $f^\prime$ drawn from the GP with kernel $k_{\hat\theta_t^{RA-BO}}(x,x^\prime)$, there exist positive constants $L_{t}, a_{t}, b_{t}$ that $\text{Pr}\{ \sup_{x\in \mathcal{X}} \vert \partial f^\prime/\partial x_h \vert > L_{t}\} \leq a_{t}e^{-(L_{t}/b_{t})^2}, h=1,\dots,d$.


Therefore, with probability greater than $(1 - dae^{-(L/b)^2} - pe^{-(C_l/q)^2} - p^\prime e^{-(C_{l^\prime}/q^\prime)^2}) \prod_{i=T_1}^t(1 - i^{1/2}(1-1/i^{1/2}) ^{\lfloor (i-1)/2 \rfloor}) \prod_{i=T_1}^t (1 - i^{1/2}(1-1/i^{1/2}) ^{M_i})$, there exists $T_2 (T_2 \geq T_1)$ that $\forall t>T_2$, the estimate $\hat{\theta}_t^{\text{\acrshort{RDEXP3}}}$ will satisfy the requirements that for all sample paths $f'$ drawn from the GP with kernel $k_{\hat{\theta}_t^{\text{\acrshort{RDEXP3}}}}(x,x')$, there exist positive constants $L_{t}, a_{t}, b_{t}$ that $\text{Pr}\{ \sup_{\bx\in \mathcal{X}} \vert \partial f'/\partial x_h \vert > L_{t}\} \leq a_{t}e^{-(L_{t}/b_{t})^2}, h=1,\dots,d$.


\end{proof}

\subsubsection{Proving Bullet Point 2 of Theorem \ref{theorem:regret-random}}

\begin{proof}

We use a similar idea as in Theorem 1 of \cite{Berkenkamp2019UnknownHyper} and the theorems and lemmas in \cite{Srinivas_2010Gaussian}. 

Given $\delta \in (0,1)$, and given the values $L_t, b_t, a_t$ defined as in Section \ref{sec:app-lipschitzt}, if we set $\beta_t = 2\log \bigl((t/2)^22\pi^2/(3\delta) \bigr) + 2d\log \bigl((t/2)^2 db_{t}r\sqrt{\log(4da_{t}/\delta)} \bigr)$, then using Lemmas 5.6, 5.7, and 5.8 in Srinivas et al, we have that, with probability greater than $(1 - dae^{-(L/b)^2} - pe^{-(C_l/q)^2} - p^\prime e^{-(C_{l^\prime}/q^\prime)^2} - p^{\prime\prime} e^{-(C_{l^{\prime\prime}}/q^{\prime\prime})^2}) - \delta) \prod_{i=T_1}^t(1 - i^{1/2}(1-1/i^{1/2}) ^{\lfloor (i-1)/2 \rfloor}) \prod_{i=T_1}^t (1 - i^{1/2}(1-1/i^{1/2}) ^{M_i})$, $\forall x \in \mathcal{X}$, and $\forall t \in 2\mathbb{N}$, $t > T_2$,  (the iteration where the data point $x_t^{RA-BO}$ in RA-BO is chosen by the GP-UCB acquisition function),
\begin{equation} \nonumber
\begin{aligned}
    & \mu_{t-1}(x_t^{RA-BO}) + \beta_t^{1/2} \sigma_{t-1}(x^{RA-BO}_t) \\
    & \geq \mu_{t-1}(x^*) + \beta_t^{1/2} \sigma_{t-1}(x^*) \geq f_{\hat{\theta}_T^{RA-BO}}(x^*) - 4/t^2,
\end{aligned}
\end{equation}
where $x^*$ denotes the objective function's global optimum, $\hat\theta_t^{RA-BO}$ is the estimated hyperparameter at iteration $t$ by RA-BO. Here $f_{\hat\theta_T^{RA-BO}}$ denotes a sample path of the GP with kernel $k_{\hat\theta_t^{RA-BO}}(x,x^\prime)$, $f_{\hat\theta_T^{RA-BO}}$ shares the observed data $D_{t-1}$ with $f$, and $f_{\hat\theta_T^{RA-BO}}$ has an upper bounded Lipschitz constant when $f$ has an upper bounded Lipschitz constant (as proven in the above paragraph, there exists such a sample path due to the closeness of $\hat\theta_t^{RA-BO}$ and $\theta^*$ and the differentiable and stationary property of the kernel $k$).

In the above inequality, we have the probability $(1 - dae^{-(L/b)^2} - pe^{-(C_l/q)^2} - p^\prime e^{-(C_{l^\prime}/q^\prime)^2} - p^{\prime\prime} e^{-(C_{l^{\prime\prime}}/q^{\prime\prime})^2}) - \delta) \prod_{i=T_1}^t(1 - i^{1/2}(1-1/i^{1/2}) ^{\lfloor (i-1)/2 \rfloor}) \prod_{i=T_1}^t (1 - i^{1/2}(1-1/i^{1/2}) ^{M_i})$ as we need to make sure we satisfy all the assumptions regarding the Lipschitz continuity of $f, l, \partial l/\partial \theta, \partial^2 l / \partial \theta^2$ and the upper bound of GP-UCB to contain the sample path $f_{\hat{\theta}_T^{RA-BO}}$ for all $t$. Again, we use the Fr\'{e}chet inequality to arrive at the first component of this probability, then multiply with the second component to get the final probability.

Now we will bound the difference between $f_{\hat\theta_T^{RA-BO}}(x^*)$ and $f(x^*)$. First, using the proof in Proposition \ref{prop:newloss}, we have that with probability $\prod_{i=T_1}^t(1 - i^{1/2}(1-1/i^{1/2}) ^{\lfloor (i-1)/2 \rfloor})$, the distance between $x^*$ and the closest data point $\hat{x}^*$ in the observed dataset $D_{t-1}$ is of $\mathcal{O}(t^{-1/(2d)})$ for all $t$. Then for all $t>T_2$, with probability larger than $\big( 1-dae^{-L^2/b^2} \big)$, we have that,

\begin{equation}
\begin{aligned}
   & \vert f_{\hat\theta_t^{\text{RA-BO}}}(x^*) - f(x^*) \vert \\
   & \quad = \vert f_{\hat\theta_t^{\text{RA-BO}}}(x^*) - f_{\hat\theta_t^{\text{RA-BO}}} (\hat{x}^*) + f(\hat{x}^*) - f(x^*) \vert \\
   & \quad \leq \mathcal{O} (\Vert x^* - \hat{x}^* \Vert_1).
\end{aligned}
\end{equation}

This inequality is due to the definition of $f_{\hat\theta_T^{RA-BO}}$: it has an upper bounded Lipschitz constant when $f$ has an upper bounded Lipschitz constant.

Combining this inequality with Assumption \ref{assum:f-smoothness}, we have that, with probability at least $\big( 1-dae^{-L^2/b} \big) \prod_{i=T_1}^t(1 - i^{1/2}(1-1/i^{1/2}) ^{\lfloor (i-1)/2 \rfloor})$, for all $t>T_2$, 
$  \vert f_{\hat\theta_t^{\text{RA-BO}}}(x^*) - f(x^*) \vert \leq \mathcal{O} (\Vert x^* - \hat{x}^* \Vert_1) \leq \mathcal{O} (t^{-1/(2d)}). $

Now with this rate, we derive the sub-linear rate of the regret $S^{\text{RA-BO}}_e(T)$ in Theorem \ref{theorem:regret-random}. Basically, we will have that, with probability at least $(1 - dae^{-(L/b)^2} - pe^{-(C_l/q)^2} - p^\prime e^{-(C_{l^\prime}/q^\prime)^2} - p^{\prime\prime} e^{-(C_{l^{\prime\prime}}/q^{\prime\prime})^2} - \delta) \prod_{i=T_1}^T(1 - i^{1/2}(1-1/i^{1/2}) ^{\lfloor (i-1)/2 \rfloor}) \prod_{i=T_1}^T (1 - i^{1/2}(1-1/i^{1/2}) ^{M_i})$, for $\forall T > T_2$,
\begin{equation} \nonumber
    \begin{aligned}
        T/2 f(x^*) - \sum_{t=2, t\in 2\mathbb{N}}^T f(x^{\text{RA-BO}}_t) \leq \sum_{t=2, t\in 2\mathbb{N}}^T 2\beta_t^{1/2}\sigma_{t-1}(x^{\text{RA-BO}}_t) + \mathcal{O}(T^{1-1/(2d)}).
    \end{aligned}
\end{equation}

Then we have that, with probability at least $\big( 1-dae^{-L^2/b} - pe^{-(C_l/q)^2} - p^\prime e^{-(C_{l^\prime}/q^\prime)^2} - p^{\prime\prime} e^{-(C_{l^{\prime\prime}}/q^{\prime\prime})^2} - \delta \big) \prod_{i=T_1}^T(1 - i^{1/2}(1-1/i^{1/2}) ^{\lfloor (i-1)/2 \rfloor}) \prod_{i=T_1}^T (1 - i^{1/2}(1-1/i^{1/2}) ^{M_i})$, for $T > T_2$,
\begin{equation} \nonumber
    \begin{aligned}
        S^{RA-BO}_e(T) \leq 2R^{RA-BO}_e(T)/T \leq \sqrt{C_1 \beta_T \gamma_{T/2}(\hat{\theta}_T^{\text{RA-BO}})/(T/2)} + \mathcal{O}(T^{-1/(2d)}).
    \end{aligned}
\end{equation}

Note that regarding the probability $\big( 1-dae^{-L^2/b} - pe^{-(C_l/q)^2} - p^\prime e^{-(C_{l^\prime}/q^\prime)^2} - p^{\prime\prime} e^{-(C_{l^{\prime\prime}}/q^{\prime\prime})^2} - \delta \big) \prod_{i=T_1}^T(1 - i^{1/2}(1-1/i^{1/2}) ^{\lfloor (i-1)/2 \rfloor}) \prod_{i=T_1}^T (1 - i^{1/2}(1-1/i^{1/2}) ^{M_i})$, like we mentioned before, we can always control the values $L, C_l, C_{l^\prime}, C_{l^{\prime\prime}}$ to make sure the values $dae^{-L^2/b}, pe^{-(C_l/q)^2}, p^\prime e^{-(C_{l^\prime}/q^\prime)^2}, p^{\prime\prime} e^{-(C_{l^{\prime\prime}}/q^{\prime\prime})^2}$ small. We can obviously choose $\delta$ small, therefore, the probability $\big( 1-dae^{-L^2/b} - pe^{-(C_l/q)^2} - p^\prime e^{-(C_{l^\prime}/q^\prime)^2} - p^{\prime\prime} e^{-(C_{l^{\prime\prime}}/q^{\prime\prime})^2} - \delta \big)$ is high. And when $T$ is large, the probability $\prod_{i=T_1}^T(1 - i^{1/2}(1-1/i^{1/2}) ^{\lfloor (i-1)/2 \rfloor}) \prod_{i=T_1}^T (1 - i^{1/2}(1-1/i^{1/2}) ^{M_i})$ is close to 1. Therefore, the probability $\big( 1-dae^{-L^2/b} - pe^{-(C_l/q)^2} - p^\prime e^{-(C_{l^\prime}/q^\prime)^2} - p^{\prime\prime} e^{-(C_{l^{\prime\prime}}/q^{\prime\prime})^2} - \delta \big) \prod_{i=T_1}^T(1 - i^{1/2}(1-1/i^{1/2}) ^{\lfloor (i-1)/2 \rfloor}) \prod_{i=T_1}^T (1 - i^{1/2}(1-1/i^{1/2}) ^{M_i})$ is a high probability.

\end{proof}

\subsection{Proof of Proposition \ref{prop:informationgain}}

\begin{proof}

The proof of this proposition is based on the proof of Proposition 2 in \cite{Berkenkamp2019UnknownHyper}. Specifically, we use the same proof technique, the only difference is that we replace the formula of $\hat{\theta}^{\text{\acrshort{RDEXP3}}}_t$ ($t>T_1$) by $\hat{\theta}^{\text{\acrshort{RDEXP3}}}_t = \theta^* + \kappa_t t^{-1/2}$, and $\kappa_t$ is upper bounded.

We first derive an upper bound on the maximum information gain corresponding to the SE kernel. To derive this upper bound, the main idea is to compute the operator spectrum of this kernel w.r.t. the uniform distribution (Theorem 8 in \cite{Srinivas_2010Gaussian}). For the SE kernel, we can have a bound on its eigenspectrum as follows \cite{Seeger_2008Information, Srinivas_2010Gaussian, Berkenkamp2019UnknownHyper},
\begin{equation}
\begin{aligned}
    & \lambda_s \leq cB^{s^{1/d}}, c=\sqrt{2a/A}, \\
    & b=1/(2(\hat{\theta}^{\text{\acrshort{RDEXP3}}}_t)^2), B=b/a, A=a+b+\sqrt{a^2+2ab}.
\end{aligned}
\end{equation}

Using the same arguments as in \cite{Srinivas_2010Gaussian, Berkenkamp2019UnknownHyper}, the constant $a>0$ parameterizes the distribution $\mu(x) \sim \mathcal{N}(0, (4a)^{-1}I_d)$. Given that $\hat{\theta}^{\text{\acrshort{RDEXP3}}}_t>0$ (as it is close to $\theta^*$ with large $t$ and for SE kernel, $\theta^*$ is positive), we have that $b\geq 0, 0<B<1, c>0$, and $A>0$. We can then bound the term $B_k(T_*)$ in the upper bound of maximum information gain used in \cite{Srinivas_2010Gaussian, Berkenkamp2019UnknownHyper} using the same technique as in \cite{Seeger_2008Information, Berkenkamp2019UnknownHyper},
\begin{equation}
\begin{aligned}
    B_k(T_*) \leq cd\alpha^{-d} (d-1)! \exp^{-\beta} \sum_{k=1}^{d-1} \beta^k/k! = c (d!) \alpha^{-d} \exp ^{-\beta} \sum_{k=1}^{d-1} (k!)^{-1} \beta^k.
\end{aligned}
\end{equation}

Similar to \cite{Srinivas_2010Gaussian, Berkenkamp2019UnknownHyper}, we assume without loss of generality that $k(x,x')=1$ for all $x \in \mathcal{X}$. Then we focus on determining how $\alpha^{-d}$ and $c$ depend on $\hat{\theta}^{\text{\acrshort{RDEXP3}}}_t$. We can quantify this through the relationship $\hat{\theta}^{\text{\acrshort{RDEXP3}}}_t = \theta^* + \kappa_t t^{-1/2}$, i.e.,
\begin{equation}
\begin{aligned}
    \alpha^{-d} &= \log^{-d}(1/B) = \log^{-d} (2(\hat{\theta}^{\text{\acrshort{RDEXP3}}}_t)^2 A) \\
    &= \log^{-d} (1+2(\hat{\theta}^{\text{\acrshort{RDEXP3}}}_t)^2a + 2\hat{\theta}^{\text{\acrshort{RDEXP3}}}_t\sqrt{a^2+a/(\hat{\theta}^{\text{\acrshort{RDEXP3}}}_t)^2)} \\
    & \leq \log^{-d} (1+2(\hat{\theta}^{\text{\acrshort{RDEXP3}}}_t)^2 a) \\
    & \leq ( \dfrac{\log (1 + 2(\theta^* + \kappa_{\max})^2a)}{ 2(\theta^* + \kappa_{\max})^2a } 2(\hat{\theta}^{\text{\acrshort{RDEXP3}}}_t)^2 a )^{-d} \\
    & = \mathcal{O}((\hat{\theta}^{\text{\acrshort{RDEXP3}}}_t) ^ {-2d}) = \mathcal{O} (\theta^* + \kappa_t t^{-1/2})^{-2d},
\end{aligned}
\end{equation}   
where we have use the inequality that for all $x \in [0, x_{\max}^2]$, it holds that $\log (1+x^2) \geq \log (1+x_{\max}^2)x^2/x_{\max}^2 $ and $\kappa_{\max}$ denotes the upper bound on $\kappa_t$. Similarly, following \cite{Berkenkamp2019UnknownHyper}, we have that,
\begin{equation}
\begin{aligned}
    c = (\dfrac{2a}{a + \dfrac{1}{2(\hat{\theta}^{\text{\acrshort{RDEXP3}}}_t)^2} + \sqrt{a^2 + \dfrac{a}{(\hat{\theta}^{\text{\acrshort{RDEXP3}}}_t)^2}}})^{d/2} \leq (\dfrac{2a}{\dfrac{1}{2(\hat{\theta}^{\text{\acrshort{RDEXP3}}}_t)^2}}) &= (4a(\hat{\theta}^{\text{\acrshort{RDEXP3}}}_t)^2)^{d/2} \\
    &= (4a (\theta^* + \kappa_t t^{-1/2})^2)^{d/2} \\
    &= \mathcal{O} (\theta^* + \kappa_t t^{-1/2})^{d}.
\end{aligned}
\end{equation}

As in \citet{Srinivas_2010Gaussian, Berkenkamp2019UnknownHyper}, we can choose $T_*=\bigl(\log(Tn_T)/\alpha \bigr)^d$, so that $\beta=\log(Tn_T)$. Plugging into the upper bound of maximum information in Theorem 8 of \cite{Srinivas_2010Gaussian}, the first term of this upper bound dominates, and therefore, for $T>T_1$, with probability $(1 - dae^{-(L/b)^2} - pe^{-(C_l/q)^2} - p^\prime e^{-(C_{l^\prime}/q^\prime)^2}) \prod_{i=T_1}^T(1 - i^{1/2}(1-1/i^{1/2}) ^{\lfloor (i-1)/2 \rfloor}) \prod_{i=T_1}^T (1 - i^{1/2}(1-1/i^{1/2}) ^{M_i})$, we have,
\begin{equation}
\begin{aligned}
    & \gamma_{T/2}(\hat{\theta}^{\text{\acrshort{RDEXP3}}}_t)
    = \mathcal{O} \bigl( (\log((T/2)^{d+1} (\log (T/2))))^{d+1} c \alpha^{-d} \bigr) \\
    &= \mathcal{O} \bigl((\log((T/2)^{d+1} (\log (T/2))))^{d+1} (\theta^* + \kappa_T T^{-1/2})^{-d} \bigr) \\
    &= \mathcal{O} ( (\log (T/2))^{d+1}).
\end{aligned}
\end{equation}


For the Mat\'{e}rn kernel with the roughness parameter $\nu$, we also use the same proof technique as in the proof of Proposition 2 in \cite{Berkenkamp2019UnknownHyper}, and by using the relationship of $\hat{\theta}^{\text{\acrshort{RDEXP3}}}_t = \theta^* + \kappa_t t^{-1/2}$ where $\kappa_t$ is upper bounded. To derive an upper bound on the maximum information gain for the Mat\'{e}rn kernel, we need to derive an upper bound on the parameter $C=C_3^{(2\nu+d)/d}$, which can be computed as follows \cite{Berkenkamp2019UnknownHyper},
\begin{equation}
\begin{aligned}
    C_t(\alpha, \nu) &= \frac{\Gamma (\nu + d/2)}{\pi^{d/2}\Gamma(\nu)} \alpha^d \\
    &= \frac{\Gamma (\nu + d/2)}{\pi^{d/2}\Gamma(\nu)} (\frac{2\pi (\theta^* + \kappa_t t^{-1/2})}{\sqrt{\nu}})^d \\
    &= \mathcal{O}((\theta^* + \kappa_t t^{-1/2})^d), \\
    c_1 &= \frac{1}{(2\pi)^d C_t(\alpha, \nu)} = \mathcal{O}((\theta^* + \kappa_t t^{-1/2})^{-d}), \\
    C_2 &= \frac{\alpha^{-d}}{ 2^d \pi^{d/2} \Gamma(d/2)} = \mathcal{O}((\theta^* + \kappa_t t^{-1/2})^{-d}), \\
    C_3 &= C_2 \frac{2\Tilde{C}}{d} c_1^{\frac{-d}{2\nu+d}} \\
    &= \mathcal{O}((\theta^* + \kappa_t t^{-1/2})^{-d} (\theta^* + \kappa_t t^{-1/2})^{-d\times\frac{-d}{2\nu+d}}) \\
    &= \mathcal{O} ((\theta^* + \kappa_t t^{-1/2})^{-d + \frac{d^2}{2\nu+d}}) \\
    &= \mathcal{O} (\theta^* + \kappa_t t^{-1/2})^{\frac{-2\nu d}{2\nu+d}}.
\end{aligned}    
\end{equation}
Now, the constant $C$ can then be upper bounded by $\mathcal{O}(\theta^* + \kappa_t t^{-1/2})^{-2\nu}$. From this upper bound, we then have that $B_k(T_*)=\mathcal{O}((\theta^* + \kappa_T T^{-1/2})^{\frac{-2\nu d}{2\nu+d}} T_*^{1-(2\nu+d)/d})$, and as in \cite{Srinivas_2010Gaussian}, we have that for $T>T_1$, with probability $(1 - dae^{-(L/b)^2} - pe^{-(C_l/q)^2} - p^\prime e^{-(C_{l^\prime}/q^\prime)^2}) \prod_{i=T_1}^T(1 - i^{1/2}(1-1/i^{1/2}) ^{\lfloor (i-1)/2 \rfloor}) \prod_{i=T_1}^T (1 - i^{1/2}(1-1/i^{1/2}) ^{M_i})$, $\gamma_{T/2}(\hat{\theta}^{\text{\acrshort{RDEXP3}}}_t) = \mathcal{O}((T/2)^{\frac{d(d+1)}{2\nu+d(d+1)}}(\log (T/2)) (\theta^* + \kappa_t T^{-1/2})^{-2\nu} ) = \mathcal{O} ((T/2)^{\frac{d(d+1)}{2\nu+d(d+1)}}(\log (T/2)))$.
\end{proof}


\subsection{Proof of Theorem \ref{theorem:final-convergence}}

\begin{proof}

To prove this theorem, we use a similar proof technique as in \cite{gopakumar2018algorithmic_NIPS}. With $r_t^{a_t}$ as the reward obtained at iteration $t$ by \acrshort{UHE}. Following the theoretical analysis of EXP3 in \cite{Auer2002nonstochastic, gopakumar2018algorithmic_NIPS}, with $\gamma_t$ being chosen as $\sqrt{4 \log 2/ \bigl((e-1)T \bigr)}$, we have that, 
\begin{equation} \label{eq:app-exp3-1}
    \max_{m\in \{1,2\} } \sum_{t=2, t\in 2\mathbb{N}}^T r_t^m - \mathbb{E} [ \sum_{t=2, t\in 2\mathbb{N}}^T r_t^{a_t}] \leq \mathcal{O} (\sqrt{T\log 2}).
\end{equation}


 Let us consider large iterations $t$s. from Theorem \ref{theorem:regret-random} and Proposition \ref{prop:informationgain}, we already have that for common kernels like Square Exponential, or Mat\'{e}rn, \acrshort{RDEXP3} converges to the global optimum for even iterations $t$ with probability $>(1 - dae^{-(L/b)^2} - pe^{-(C_l/q)^2} - p^\prime e^{-(C_{l^\prime}/q^\prime)^2} - p^{\prime\prime} e^{-(C_{l^{\prime\prime}}/q^{\prime\prime})^2}) \prod_{i=T_2}^T (1 - i^{1/2}(1-1/i^{1/2}) ^{\lfloor (i-1)/2 \rfloor})(1 - i^{1/2}(1-1/i^{1/2}) ^{M_i})$ with $T_2$ defined in Theorem \ref{theorem:regret-random}. This means $\max_{m \in \{1,2\}} \max(f_t^m, f_{t-1}^m)$ also converges to the global optimum. 
 
 This means $\max(y_t, y_{t-1})$ converges to $\max(f(x_t), f(x_{t-1})) + \max(\epsilon_t, \epsilon_{t-1})$ for large iterations. Note that the mean of the random variable $\max(\epsilon_t, \epsilon_{t-1})$ is a constant, therefore, dividing two sides of Eq. (\ref{eq:app-exp3-1}) to $T$, we have that $\max(f_t^{a_t}, f_{t-1}^{a_t})$ also converges to the global optimum. And thus, from here, using the property of $\max(y_t, y_{t-1})$ converging to $\max(f(x_t), f(x_{t-1})) + \max(\epsilon_t, \epsilon_{t-1})$, for Square Exponential, or Mat\'{e}rn kernel, with probability $>(1 - dae^{-(L/b)^2} - pe^{-(C_l/q)^2} - p^\prime e^{-(C_{l^\prime}/q^\prime)^2} - p^{\prime\prime} e^{-(C_{l^{\prime\prime}}/q^{\prime\prime})^2}) \prod_{i=T_2}^T (1 - i^{1/2}(1-1/i^{1/2}) ^{\lfloor (i-1)/2 \rfloor})(1 - i^{1/2}(1-1/i^{1/2}) ^{M_i})$, Eq. (\ref{eq:app-exp3-1}) can be written as,
\begin{equation} \label{eq:app-exp3}
    \max_{m\in \{1,2\} } \sum_{t=2, t\in 2\mathbb{N}}^T f(x_t^m) - \mathbb{E} [ \sum_{t=2, t\in 2\mathbb{N}}^T f(x_t)] \leq \mathcal{O} (\sqrt{T\log 2}),
\end{equation}
where $x_t^m$ denotes the data point chosen by arm $m$ at iteration $t$, and $x_t$ denotes the data point chosen by \acrshort{UHE} at iteration $t$.

We then use our results developed in Theorem \ref{theorem:regret-random}. Specifically, we have that, for the first arm, the cumulative regret at even iterations $R^{\acrshort{RDEXP3}}_e(T)$, with probability larger than $(1 - dae^{-(L/b)^2} - pe^{-(C_l/q)^2} - p^\prime e^{-(C_{l^\prime}/q^\prime)^2} - p^{\prime\prime} e^{-(C_{l^{\prime\prime}}/q^{\prime\prime})^2}) \prod_{i=T_2}^T (1 - i^{1/2}(1-1/i^{1/2}) ^{\lfloor (i-1)/2 \rfloor})(1 - i^{1/2}(1-1/i^{1/2}) ^{M_i})$, is upper bounded by $\sqrt{C_1\beta_T\gamma_{T/2}(\hat{\theta}^{\text{\acrshort{RDEXP3}}}_t)} + \text{const}$, when $T>T_2$. This means, when $T>T_2$, with probability larger than $(1 - dae^{-(L/b)^2} - pe^{-(C_l/q)^2} - p^\prime e^{-(C_{l^\prime}/q^\prime)^2} - p^{\prime\prime} e^{-(C_{l^{\prime\prime}}/q^{\prime\prime})^2}) \prod_{i=T_2}^T (1 - i^{1/2}(1-1/i^{1/2}) ^{\lfloor (i-1)/2 \rfloor})(1 - i^{1/2}(1-1/i^{1/2}) ^{M_i})$, we have that,
\begin{equation}
\begin{aligned}
    T/2 f(x^*) - \sum_{t=2, t\in 2\mathbb{N}}^T f(x^{{\acrshort{RDEXP3}}}_t) \leq \sqrt{C_1\beta_T\gamma_{T/2}(\hat{\theta}^{\text{\acrshort{RDEXP3}}}_t)} + \text{const}.
\end{aligned}
\end{equation}

Combining this equation with Eq. (\ref{eq:app-exp3}), we have that, when $T>T_2$, with probability larger than $(1 - dae^{-(L/b)^2} - pe^{-(C_l/q)^2} - p^\prime e^{-(C_{l^\prime}/q^\prime)^2} - p^{\prime\prime} e^{-(C_{l^{\prime\prime}}/q^{\prime\prime})^2}) \prod_{i=T_2}^T (1 - i^{1/2}(1-1/i^{1/2}) ^{\lfloor (i-1)/2 \rfloor})(1 - i^{1/2}(1-1/i^{1/2}) ^{M_i})$,
\begin{equation}
\begin{aligned}
    & \max_{m\in \{1,2\} } \sum_{t=2, t\in 2\mathbb{N}}^T f(x_t^m) - \mathbb{E} [ \sum_{t=2, t\in 2\mathbb{N}}^T f(x_t)] + T/2 f(x^*) - \sum_{t=2, t\in 2\mathbb{N}}^T f(x^{{\acrshort{RDEXP3}}}_t) \\
    & \leq \sqrt{C_1\beta_T\gamma_{T/2}(\hat{\theta}^{\text{\acrshort{RDEXP3}}}_t)} + \mathcal{O} (\sqrt{T\log 2}) + \text{const}.
\end{aligned}
\end{equation}

Since $\max_{m\in \{1,2\} } \sum_{t=2, t\in 2\mathbb{N}}^T f(x_t^m) \geq \sum_{t=2, t\in 2\mathbb{N}}^T f(x^{{\acrshort{RDEXP3}}}_t)$, we have, when $T>T_2$, with probability larger than $(1 - dae^{-(L/b)^2} - pe^{-(C_l/q)^2} - p^\prime e^{-(C_{l^\prime}/q^\prime)^2} - p^{\prime\prime} e^{-(C_{l^{\prime\prime}}/q^{\prime\prime})^2}) \prod_{i=T_2}^T (1 - i^{1/2}(1-1/i^{1/2}) ^{\lfloor (i-1)/2 \rfloor})(1 - i^{1/2}(1-1/i^{1/2}) ^{M_i})$,
\begin{equation}
\begin{aligned}
   T/2 f(x^*) - \mathbb{E} [ \sum_{t=2, t\in 2\mathbb{N}}^T f(x_t)] \leq  
\sqrt{C_1\beta_T\gamma_{T/2}(\hat{\theta}^{\text{\acrshort{RDEXP3}}}_t)} + \mathcal{O} (\sqrt{T\log 2}) + \text{const}.
\end{aligned}
\end{equation}
We have that, when $T>T_2$, with probability larger than $(1 - dae^{-(L/b)^2} - pe^{-(C_l/q)^2} - p^\prime e^{-(C_{l^\prime}/q^\prime)^2} - p^{\prime\prime} e^{-(C_{l^{\prime\prime}}/q^{\prime\prime})^2}) \prod_{i=T_2}^T (1 - i^{1/2}(1-1/i^{1/2}) ^{\lfloor (i-1)/2 \rfloor})(1 - i^{1/2}(1-1/i^{1/2}) ^{M_i})$,
\begin{equation}
\begin{aligned}
    f(x^*) - \mathbb{E} \bigl[\max_{t=2,\dots,T, t\in 2\mathbb{N}} f(x_t) \bigr] & \leq f(x^*) - 2/T \mathbb{E} [ \sum_{t=2, t\in 2\mathbb{N}}^T f(x_t)] \\
    & \leq \mathcal{O} (\sqrt{2\beta_T\gamma_{T/2}(\hat{\theta}^{\text{\acrshort{RDEXP3}}}_t)/T} + \mathcal{O} (\sqrt{2 \log 2/T}).
\end{aligned}
\end{equation}

Finally, let us consider the expected simple regret, i.e., $\mathbb{E}[S_T] = f(x^*) - \mathbb{E} \bigl[\max_{t=1,\dots,T} f(x_t) \bigr]$, we have that, when $T>T_2$, with probability larger than $(1 - dae^{-(L/b)^2} - pe^{-(C_l/q)^2} - p^\prime e^{-(C_{l^\prime}/q^\prime)^2} - p^{\prime\prime} e^{-(C_{l^{\prime\prime}}/q^{\prime\prime})^2}) \prod_{i=T_2}^T (1 - i^{1/2}(1-1/i^{1/2}) ^{\lfloor (i-1)/2 \rfloor})(1 - i^{1/2}(1-1/i^{1/2}) ^{M_i})$,
\begin{equation}
\begin{aligned}
    f(x^*) - \mathbb{E} \bigl[\max_{t=1,\dots,T} f(x_t) \bigr] &\leq f(x^*) - \mathbb{E} \bigl[\max_{t=2,\dots,T, t\in 2\mathbb{N}} f(x_t) \bigr] \\
    &\leq \mathcal{O} (\sqrt{2\beta_T\gamma_{T/2}(\hat{\theta}^{\text{\acrshort{RDEXP3}}}_t)/T} + \mathcal{O} (\sqrt{2 \log 2/T}).
\end{aligned}
\end{equation}

Substitute the upper bounds on the information gains obtained from Proposition \ref{prop:informationgain} into this inequality, we prove the theorem.
\end{proof}

\section{Experiment Setup} \label{appendix_sec:exp_setup}

\paragraph{Experimental Setup.} All experimental results are repeated over $10$ independent runs with different random seeds. We use Mat\'{e}rn 5/2 ARD kernels for the GP in all methods \cite{Rasmussen_2006gaussian}. For the MCMC approach, we generate $200$ burn-in and $10$ real MCMC samples. All experiments are performed on a Window workstation with Intel Xeon Silver 4208 2.1GHz 64GB RAM. For GP-UCB based approaches, following the standard practice, we set the exploration-exploitation parameter to be a constant, i.e. $\beta_t=1.96$ \citep{Berkenkamp2019UnknownHyper}. We set the parameter $M_t = 2\vert D_{t-1} \vert$.

\paragraph{Baselines.} We compare our proposed method \acrshort{UHE} against six state-of-the-art methods for estimating hyperparameters for GP in BO: (1) \textbf{Maximum A Posteriori (MAP)} method; (2) \textbf{Markov Chain Monte Carlo (\acrshort{MCMC})} via HMC \citep{Neal2011HMC}; (3) \textbf{A-GP-UCB}: adaptive GP-UCB for BO with unknown hyperparameters \citep{Berkenkamp2019UnknownHyper}; (4) \textbf{Wang \& de Freitas' method}: BO with unknown hyperparameters proposed in \citep{Wang_2014Theoretical}; (5) \textbf{Portfolio}: described in \citet{Hoffman2011Portfolio} with the choice of acquisition functions being random and GP-UCB; (6) \textbf{Random}: the data points are chosen randomly. We are unable to compare with \citet{BogunovicK21}, which also tackles the problem of BO with unknown GP hyperparameters. There were no empirical experiments conducted in \cite{BogunovicK21} so it is unclear how the methods can be implemented and empirically evaluated.

The detailed implementations of all the methods are as follows.
\begin{enumerate}
    \item \textbf{\acrshort{UHE}}: We use the official document of GPy\footnote{\url{https://nbviewer.jupyter.org/github/SheffieldML/notebook/blob/master/GPy/index.ipynb}} and the source code in \citep{Lyu2018BatchBOMultiAcq} to implement \acrshort{UHE} . The prior for each type of GP hyperparameters is set as follows. The prior for the kernel variance is set as a Gamma($10^{-3}$, $10$), the prior for the noise variance is set as Gamma($10^{-3}$, $10$), and the prior for the kernel lengthscales is set as Gamma($10^{-3}$, $10$).
    \item \textbf{MAP}: The source code and all settings for the GPs (e.g. priors types, priors hyperparameters) are the same as those in \acrshort{UHE} .
    \item \textbf{MCMC}: The source code and all settings for the GPs (e.g. priors types, priors hyperparameters, HMC samples) are the same as those in \acrshort{UHE} .
    \item \textbf{A-GP-UCB} \citep{Berkenkamp2019UnknownHyper}: Since the source code of this method is not published, we implement the method by ourselves. All the settings for the GPs (e.g. priors types, priors hyperparameters)  are the same as those in \acrshort{UHE}. We set the sub-linear reference function $p(t)$ to be $t^{0.9}$ as recommended in the work. We use the one-step prediction technique with Eq. (16) in the paper, combined with a grid search method to compute the function $h(t)$, and then we also set $g(t)=h(t)$ as suggested in the work. Finally, we also use the MAP estimate $\hat{\theta}^{\textsc{MAP}}_t$ combined with the sub-linear reference function $p(t)$ to construct the adaptive lengthscales in each iteration.
    \item \textbf{Wang \& de Freitas' method} \citep{Wang_2014Theoretical}: Since the source code of this method is not published, we implement the method by ourselves. All the settings for the GPs (e.g. priors types, priors hyperparameters)  are the same as those in \acrshort{UHE} . We use the same variant of the method that was compared in \citep{Berkenkamp2019UnknownHyper}, that is, we conduct a grid search to find the smallest possible scaling factor that renders $\sigma_t(x_{t+1}) \geq \kappa=0.1$.
    \item \textbf{Porfolio} \citep{Hoffman2011Portfolio}: Since the source code of this method is not published, we implement the method by ourselves. All the settings for the GPs (e.g. priors types, priors hyperparameters)  are the same as those in \acrshort{UHE}. In this method, the porfolio of acquisition functions consists of random and the GP-UCB acquisition function. EXP3 method is used to select the acquisition function (random or GP-UCB) in each iteration.
    \item \textbf{Random}: The data points are chosen randomly within the search domain $\mathcal{X}$.
\end{enumerate}

\paragraph{Synthetic Benchmarks.} We consider four functions: Branin, Deceptive, Hartmann3, and h1. The descriptions of these synthetic functions are as follows.
\begin{itemize}
    \item \textbf{Branin}: The formula of the Brannin function is: $f(x) = a(x_2 - bx_1^2 + cx_1 - r)^2 + s(1-t)\cos(x_1)+s$ with $a=1, b=5.1/(4\pi^2),c=5/\pi, c=5/\pi,r=6,s=10,t=1/(8\pi)$ \footnote{\label{sfu_site}\url{https://www.sfu.ca/~ssurjano/optimization.html}}.
    \item \textbf{Hartmann3}: The formula of the Hartmann3 function is: $f(x) = -\sum_{i=1}^4 \alpha_i \exp(-\sum_{j=1}^3 A_{ij}(x_j-P_{ij})^2)$ where $\alpha=[1.0, 1.2, 3.0, 3.2]^T$, $A = [3.0\ 10\ 30, 0.1\ 10\ 35, 3.0\ 10\ 30, 0.1\ 10\ 35]$, and $P=10^{-4} [3689\ 1170\ 2673, 4699\ 4387\ 7470, 1091\ 8732\ 55 \\ 
    47, 381\ 5743\ 8828]$\footref{sfu_site}.
    \item \textbf{Deceptive}: The formula of the Deceptive function is $f(x) = - (1/n \sum_{i=1}^n g_i (x_i))^{\beta}$ with $n$ representing the number of dimensions, $\beta=2$, and $g(.)$ is defined as\footnote{\url{http://infinity77.net/global\_optimization/test\_functions\_nd\_D.html}},
    \begin{equation} \nonumber
        g_i(x_i) = \begin{cases} - \frac{x}{\alpha_i} + \frac{4}{5} & \textrm{if} \hspace{5pt} 0 \leq x_i \leq \frac{4}{5} \alpha_i \\ \frac{5x} {\alpha_i} -4 & \textrm{if} \hspace{5pt} \frac{4}{5} \alpha_i \le x_i \leq \alpha_i \\ \frac{5(x - \alpha_i)}{\alpha_i-1} & \textrm{if} \hspace{5pt} \alpha_i \le x_i \leq \frac{1 + 4\alpha_i}{5} \\ \frac{x - 1}{1 - \alpha_i} & \textrm{if} \hspace{5pt} \frac{1 + 4\alpha_i}{5} \le x_i \leq 1 \end{cases},
    \end{equation}
    with $\alpha_i=i/(n+1)$.
    \item \textbf{h1}: The formula of the h1 function is $f(x) = ((\sin(x_1 - x_2/8))^2 + (\sin(x_2+x_1/8))^2)/ \sqrt{(x_1 - 8.6998)^2 + (x_2 - 6.7665)^2+1}$ \cite{Soest2003h1func}.
\end{itemize}

\paragraph{Real-world Benchmarks.} Details about the real-world benchmarks that we use to evaluate \acrshort{UHE} and the baselines are as follows.
\begin{itemize}
    \item \textbf{Logistic Regression}: We use the Logistic Regression implementation from the HPOBench benchmark \cite{eggensperger2021hpobench} to perform classification on the OpenML's Vehicle Silhouettes dataset~\cite{Siebert1987VehicleRU}. This benchmark has 2 hyperparameters: learning rate and regularization. Both are tuned in an exponent space (base 10), with the search domain of $[-5, 0]$, as in the HPOBench benchmark. The dataset is split into $2/3$ for training and $1/3$ for testing. 
    \item \textbf{XGBoost}: The task is to train an XGBoost model \citep{chen2016xgboost} on the OpenML's Software Defect dataset \cite{Sayyad2005Promise}. We use the implementation from the HPOBench benchmark \cite{eggensperger2021hpobench}. This benchmark has 4 hyperparameters that tune the maximum depth per tree, the features subsampled per tree, the learning rate and the L2 regularization for the XGBoost model. The search domains of these 4 hyperparameters are set as in the HPOBench benchmark. The dataset is split into $2/3$ for training and $1/3$ for testing.
    \item \textbf{Random Forest:} We use the Random Forest implementation from the HPOBench benchmark \cite{eggensperger2021hpobench} to perform classification on the OpenML's Wilt dataset \cite{johnson2013hybrid}. This benchmark has 4 hyperparameters: max depth, min sample split, max features, min sample leaf. The search domain of these 4 hyperparameters are set as in the HPOBench benchmark. The dataset is split into $2/3$ for training and $1/3$ for testing.
\end{itemize}

\section{Ablation Study} \label{appendix_sec:add_exps}

\begin{figure*}[t]
    \centering       
    \includegraphics[trim=0cm 0cm 0cm  0cm, clip, width=0.325\textwidth]{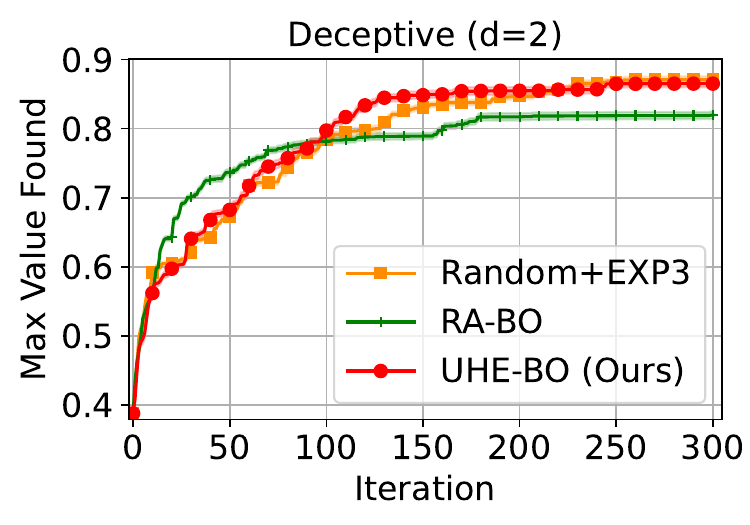}
        \includegraphics[trim=0cm 0cm 0cm  0cm, clip, width=0.325\textwidth]{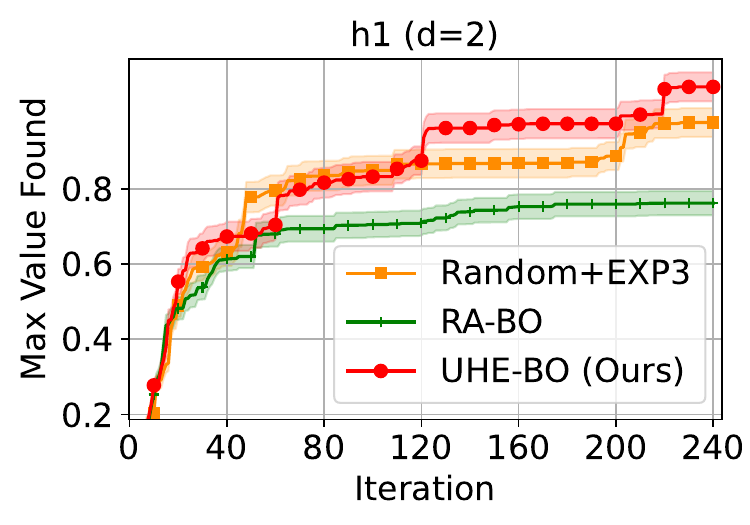}
        \includegraphics[trim=0cm 0cm 0cm  0cm, clip, width=0.325\textwidth]
        {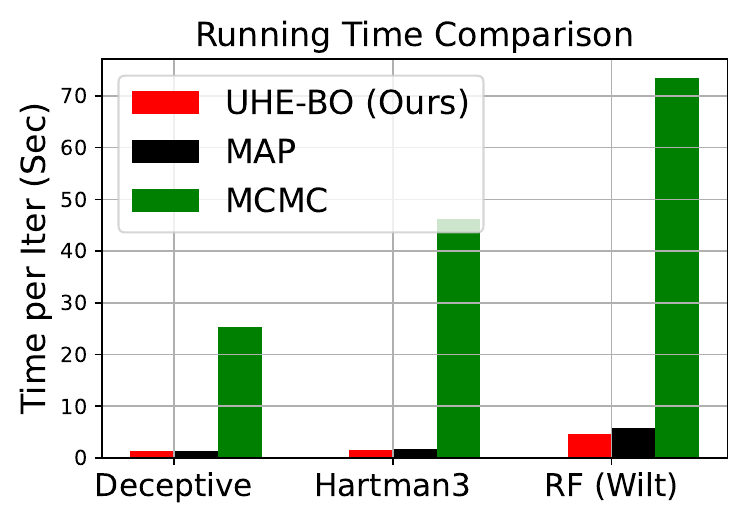}
    \vspace{-2pt}
    \caption{
    \textbf{Left} \& \textbf{Middle}: The performance of our method (\acrshort{UHE}), the method \acrshort{RDEXP3}, and the method with i.i.d. sampling via EXP3 (Random+EXP3) on two problems. \textbf{Right}: The running time of our method is comparable to MAP whilst much faster than MCMC across three problems.
    }    
    \label{fig:ablation-component}
   \vspace{-4mm}
\end{figure*}

We show in Fig. \ref{fig:ablation-component} the ablation study to understand the effect of each component in \acrshort{UHE}, as described in Sec. \ref{sec:ablation}.

\end{document}